\documentclass[10pt,letterpaper,conference]{IEEEtran}
\pdfoutput=1

\makeatletter

\usepackage{authblk}
\usepackage[ruled,vlined]{algorithm2e}
\usepackage[inkscapelatex=false]{svg}
\usepackage{multicol}
\usepackage{inputenc}
\usepackage{amsfonts}
\usepackage{amsmath}
\usepackage{amssymb}
\usepackage[T1]{fontenc}
\usepackage{microtype}
\usepackage[dvipsnames]{xcolor}
\usepackage{graphicx}
\usepackage{graphics}
\usepackage{enumitem}
\usepackage[font=footnotesize]{caption}
\usepackage{subcaption}
\usepackage{booktabs}
\usepackage{multirow}
\usepackage[export]{adjustbox}
\usepackage{breqn}
\usepackage{acronym}
\usepackage{setspace}
\usepackage{bm}
\usepackage{stackengine}
\usepackage{siunitx}
\usepackage{makecell}
\usepackage{smartdiagram}
\usepackage{tikz}
\usesmartdiagramlibrary{additions}
\usepackage{hyperref}
\definecolor{azure}     {rgb}{0,0.5,1}
\definecolor{dkpowder}  {rgb}{0,0.2,0.7}
\definecolor{deepred}   {rgb}{0.7,0,0}
\definecolor{deepblue}  {rgb}{0,0,0.7}
\definecolor{deepgreen} {rgb}{0,0.5,0}
\definecolor{deeporange}{rgb}{0.91, 0.41, 0.17}
\hypersetup{%
	pdfpagemode  = {UseOutlines},
	bookmarksopen,
	pdfstartview = {FitH},
	colorlinks,
	linkcolor = {dkpowder},
	citecolor = {dkpowder},
	urlcolor  = {dkpowder},
}
\usepackage{listings}
\renewcommand{\thetable}{\arabic{table}}

\graphicspath{{/images}}
\makeatletter
\def\endthebibliography{%
	\def\@noitemerr{\@latex@warning{Empty `thebibliography' environment}}%
	\endlist}
\makeatother
\definecolor{label-running} {RGB}{ 31,119,180}
\definecolor{label-walking} {RGB}{255,127, 14}
\definecolor{label-jumping} {RGB}{ 44,160, 44}
\definecolor{label-standing}{RGB}{148,103,189}
\definecolor{label-sitting} {RGB}{140, 86, 75}
\definecolor{label-lying}   {RGB}{127,127,127}
\definecolor{label-falling} {RGB}{188,189, 34}
\definecolor{label-transit} {RGB}{ 23,190,207}

\IEEEoverridecommandlockouts

\usepackage[backend=bibtex]{biblatex}
\usepackage{todonotes}
\usepackage{tabularx}
\usepackage{xargs}  
\usepackage{arydshln}
\usepackage{array}       
\usepackage{makecell}    
\newcommandx{\info}[2][1=]{\todo[linecolor=OliveGreen,backgroundcolor=OliveGreen!25,bordercolor=OliveGreen,#1]{#2}}

\bibliography{IEEEabrv,bib/bibliography.bib}

\title{\LARGE \bf Autonomous Docking of Multi-Rotor UAVs on Blimps under the \\Influence of Wind Gusts}
\author{Pascal Goldschmid and Aamir Ahmad 
\thanks{Both authors are with the University of Stuttgart, Faculty of Aerospace Engineering and Geodesy, Institute of Flight Mechanics and Control (iFR), Flight Robotics and Perception Group (FRPG). Pfaffenwaldring 27, 70569 Stuttgart, Germany. Aamir Ahmad is also with the Max Planck Institute for Intelligent Systems, T\"ubingen, Perceiving Systems Department. Max-Planck-Ring 4, 71069 T\"ubingen, Germany. {\tt\footnotesize 
(pascal.goldschmid,aamir.ahmad)@ifr.uni-stuttgart.de}} 
\thanks{The authors thank Eric Price, Corinna Lipp and Dominik Goldschmid for their help with the flight experiments. The authors thank the International Max Planck Research School for Intelligent Systems (IMPRS-IS) for supporting Pascal Goldschmid.}
}

\makeatother
\begin{document}
\setlength{\belowcaptionskip}{-5pt}
\maketitle

\begin{abstract}
Multi-rotor UAVs face limited flight time due to battery constraints. Autonomous docking on blimps with onboard battery recharging and data offloading offers a promising solution for extended UAV missions. However, the vulnerability of blimps to wind gusts causes trajectory deviations, requiring precise, obstacle-aware docking strategies. To this end, this work introduces two key novelties: (i) a temporal convolutional network that predicts blimp responses to wind gusts, enabling rapid gust detection and estimation of points where the wind gust effect has subsided; (ii) a model predictive controller (MPC) that leverages these predictions to compute collision-free trajectories for docking, enabled by a novel obstacle avoidance method for close-range maneuvers near the blimp. Simulation results show our method outperforms a baseline constant-velocity model of the blimp significantly across different scenarios. We further validate the approach in real-world experiments, demonstrating the first autonomous multi-rotor docking control strategy on blimps shown outside simulation. Source code is available here \url{https://github.com/robot-perception-group/multi_rotor_airship_docking}.
\\
\\
Keywords: docking, applications, mechanics and control, perception and autonomy
\end{abstract}

\section{Introduction}
Multi-rotor vehicles suffer from restricted range and flight time due to limited battery capacity. This makes autonomous docking on other (aerial) autonomous platforms an attractive solution approach for replenishing batteries and offloading data \cite{goldschmid2024}. 

So far, in comparison to autonomous fixed-wing and multi-rotor aircraft,  little research has been conducted in the field of autonomous lighter-than-air vehicles. Recent advancements in autonomous blimp control, e.g. \cite{price2022,liu2022}, now make this type of vehicle available for more complex tasks such as docking in an easy-to-use manner.
\begin{figure}[htbp]
    \centering
    \includegraphics[width=0.5\textwidth]{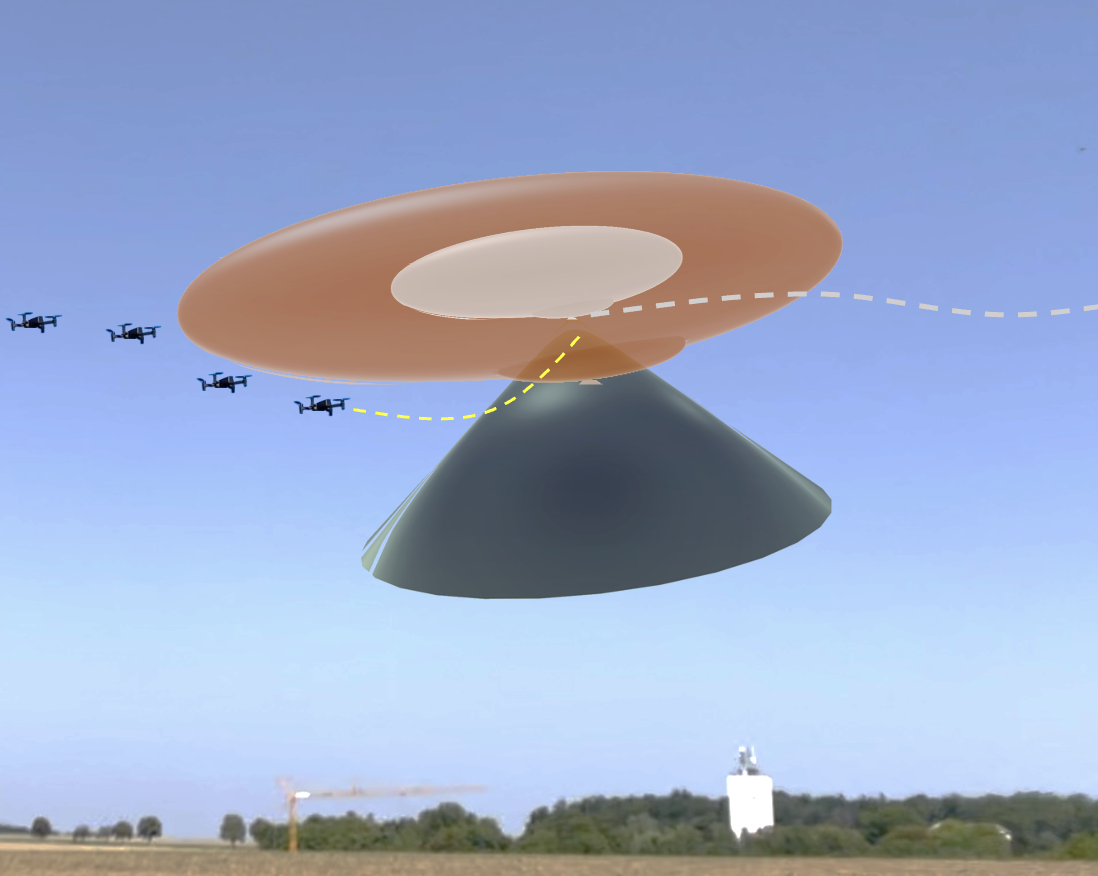}
    \caption{Illustration of our approach. A multi-rotor UAV approaches a docking port mounted on a virtual blimp (white). The blimp is protected by a no-fly zone (red) enhanced with a cone shaped approach corridor (green). A collision-free trajectory is planned (yellow line) considering the predicted trajectory of the blimp (grey line) under the influence of wind gusts.}
    \label{fig:title}
\end{figure}
Blimps, a type of lighter-than-air vehicle with a non-rigid structure, possess a number of qualities which make them suitable as flying carrier hubs where multi-rotor UAVs can dock, e.g. high payload capacity, long flight time and flight range. However, they are highly susceptible to wind gusts \cite{price2022} which can cause strong oscillations in the blimp motion which is slow to decline. Furthermore, the oscillating motion is affected by a variety of variables such as the wind gust itself (direction, strength, duration) and characteristics of the blimp (e.g. buoyancy, presence of non-rigid structure and consequently the ability to twist and bend when deflation occurs) thus making docking a dangerous venture. Consequently, autonomous multi-rotor docking on blimps poses strict requirements to the control algorithm of the multi-rotor UAV. This is due to the conflict between the necessity of maneuvering in close proximity to the blimp  and the general interest of collision avoidance.

Autonomous aerial docking between heterogeneous vehicles has been demonstrated for multi-rotor UAVs and fixed-wing aircraft \cite{caruso2021,hove2023}, but existing methods overlook wind disturbances and lack obstacle avoidance during final approach. No solution yet addresses docking on lighter-than-air platforms, where the slow dynamics of blimps \cite{price2022,liu2022} make accurate, disturbance-aware motion prediction critical. While Model Predictive Control can incorporate constraints \cite{nguyen2021,tallamraju2018}, its reliance on simplified dynamics models \cite{salzmann2023} typically limits its applicability. To address this issue, we propose a data-driven approach using Temporal Convolutional Networks (TCNs) \cite{lea2017,bai2018} to capture the long lasting blimp motion under wind gusts. Recursive forecasting, i.e. using generated output of the TCN as input for the subsequent prediction,  is used to achieve longer predictions. We provide a simulated dataset, a generation pipeline, analysis tools, and a trained TCN predictor to enable robust docking of multi-rotor UAVs on blimps.

Our overall approach is as follows. A TCN predicts the blimp’s trajectory via recursive forecasting, from which a target position for the multi-rotor UAV is derived. The multi-rotor UAV tracks this setpoint using quadratic MPC introduced in \cite{goldschmid2025}. To protect the blimp’s hull, a no-fly zone is established, except for a cone-shaped approach corridor beneath the blimp leading to the docking port at its center of gravity. During docking, the predicted trajectory is monitored using a wind gust detection method. It is based on the determination of the maximum deviation of the predicted blimp velocity from the its mean value.  Upon detecting a wind gust, the corridor closes, and the obstacle avoidance method directs the multi-rotor UAV outside the no-fly zone. Concurrently, the method identifies when gust effects subside, enabling the multi-rotor UAV to reposition and resume docking. The overall approach is illustrated by Figure \ref{fig:title}.\\
In light of this context, this work introduces the following core novelties.
\begin{enumerate}
\item \textit{Blimp response prediction:} A  framework for predicting the response of a blimp to a wind gust, including a simulated dataset, a generation pipeline, analysis tools, and a trained TCN predictor. 
\item \textit{Wind gust detection:} An approach to detect the presence of a wind gust in the blimp's predicted trajectory.
\item \textit{Collision-free trajectory planning:} We introduce the Corridor Enhanced Tangential Hull (CETH) method for obstacle avoidance. This method is embedded in an MPC-based algorithm for the planning of collision-free trajectories from the multi-rotor UAVs current position towards the docking port. Our method features a cone-shaped approach corridor within which maneuvering of the multi-rotor UAV in close proximity to the blimp is allowed. 
\item \textit{Simulation and real-world experiments:} We demonstrate our method in both simulation and a real world experiment. To the best of our knowledge, providing a control solution for wind gust aware docking of a multi-rotor UAV on a lighter-than-air vehicle is the first of its kind.
\end{enumerate}

The remainder of this work is structured as follows. Section \ref{sec:methodology} presents our developed methods, followed by \ref{sec:implementation} describing details of our implementation and experiment plan. In Section \ref{sec:experiments} we present and discuss the results of experiments before concluding this work in Section  \ref{sec:conclusion}.

\section{Related Work}
	
\subsection{Aerial Docking}
Aerial docking, often also referred to as aerial rendezvous, has been widely researched for both docking partners being of the same type. Aerial rendezvous between fixed-wing aircraft is an active topic in the context of autonomous aerial refuelling missions, e.g. in \cite{parsons2019,zhu2016,dai2018a,dai2018b,liu2019}.  All these works, as is generally common in the field, conduct experiments in simulation only. In our work we provide a  simulation framework for detailed testing  before verifying the efficacy of our approach  in real-world experiments. Unlike any of the aforementioned works, we make all simulation frameworks public.

Conducting real-world experiments is much more common for aerial docking between multi-rotor vehicles, e.g. in \cite{miyazaki2018,shankar2021,shankar2024}. A common problem in multi-rotor vertical docking is to avoid the downwash of the upper vehicle. This is addressed by a winch mechanism allowing docking outside the downwash domain \cite{miyazaki2018} or by learning the downwash force with a curriculum learning strategy \cite{shankar2024}. In our scenario, although the docking port is located in the vicinity of the blimp's actuators, they do not necessarily produce downwash which interferes with the multi-rotor vehicle's approach direction. Furthermore, since a lighter-than-air vehicle does not rely on the actuators to stay airborne, the actuators of the blimp can be disabled whenever the multi-rotor vehicle is in the last steps of the docking procedure. 

In \cite{shankar2021} a docking approach for a leader-follower scenario is presented. The leader is assumed to be on a straight line trajectory which is estimated by the follower using GPS data shared by the leader in combination with passive fiducial markers. In our work, we use a TCN to predict the trajectory of the blimp (the leader) which is also controlled to follow a straight line trajectory in the absence of wind. In the event of a wind gust, our TCN is capable of predicting the highly non-linear trajectory of the blimp during its efforts to return to the original straight line path. The multi-rotor vehicle (the follower) follows a similar approach as \cite{shankar2021} to estimate the blimp location by fusing GPS / state estimate data with readings of a fiducial marker.
Cross-plattform docking, i.e. the two UAVs involved in the docking procedure are of different types, is much less common in research, e.g. \cite{caruso2021,hove2023}. In \cite{caruso2021},   a full solution approach for multi-rotor UAV on fixed-wing aircraft is provided. A docking port is built, while guidance and control algorithms use state-of-the-art tools and hardware. 
\cite{hove2023} proposes a solution where a multi-rotor vehicle is augmented with additional propellers to control the longitudinal and lateral movement of the vehicle independently of the attitude. However, both approaches do not consider collision avoidance methods to prevent the multi-rotor vehicle to touch the fixed-wing in positions other than the docking port, e.g. in case of disturbances. In addition, the effect of wind gusts, which  can be severe for fixed-wing aircraft also as they rely on aerodynamic principles to generate lift, is neglected. In our work, we focus on developing and testing control algorithms in simulation and real-world scenarios where wind gusts and obstacle avoidance are taken into consideration. 
\subsection{Learning of Blimp Trajectories}
Deep learning based approaches have been used for path planning of stratospheric airships that are subjected to constraints such as a dynamic wind field and energy cycles \cite{qi2024}, \cite{zheng2024}. Here, wind fields are updated with low frequency (e.g. once within several hours). In \cite{qi2024}, the Soft-Actor-Critic (SAC) reinforcement learning algorithm is used to produce trajectories.  In \cite{zheng2024} a deep reinforcement learning strategy is presented that uses the D3RQN algorithm and a convolutional neural network (CNN)  to enable a stratospheric airship to reach a target point under energy constraints.  Our work however, uses supervised learning to learn the blimp's trajectory in response to a wind gust on a shorter time scale leveraging a temporal convolutional network to capture temporal dependencies.  It is common that learning based approaches are trained in simulation only. In \cite{price2022} a simulation environment and cascaded PI framework for autonomous blimp control is presented. It is well-suited for modeling the non-rigid characteristics of a blimp and has been validated in real-world experiments. We use this simulation environment to collect data for the training of the TCN.

\subsection{Temporal Convolutional Networks}
Convolutional networks have been used for time series processing e.g. in \cite{bai2018,lea2017,pantiskas2020} and with modifications to output features in \cite{gall2024}. To address the problem of the output sequence being of the same length than the input sequence we choose recursive forecasting \cite{ji2005}, an intuitive approach where generated output is reused to generate the subsequent prediction in a multi-step fashion. Even though there exist techniques to deal with the accumulating error for longer forecasting horizons, e.g. in \cite{herrera2007}, in our work, we achieve sufficient prediction accuracy without applying correction methods.

\subsection{Model Predictive Control}
Model Predictive Control  is a widely used control strategy applied accross many different research fields. Relevant to our work is \cite{tallamraju2018}, where a method is described that uses a quadractic MPC  convexifying the non-convex constraints and dependencies by replacing them as pre-computed forces in the robot dynamics. This allows for the possibility that pre-computed repulsive and tagential forces acting around the obstacle create a tangential band that can ensure collision free navigation of the multi-rotor vehicle. However, the method assumes a safety distance around obstacles which must never be penetrated. In a docking scenario where the blimp's delicate hull must be protected from the propeller blades of the multi-rotor vehicle the blimp itself is considered an obstacle which needs to be avoided except in the vicinity of the docking port. Building upon \cite{tallamraju2018}, we introduce the CETH method to open up an approach corridor towards the docking port when the external circumstances (wind gusts, accuracy of the state estimate) allow it. 

\section{Methodology}\label{sec:methodology}
\subsection{Problem Statement} \label{sec:problem_statement}
\begin{figure}[htbp]
    \centering
    \begin{subfigure}[b]{0.15\textwidth}
        \centering
        \includegraphics[width=\linewidth]{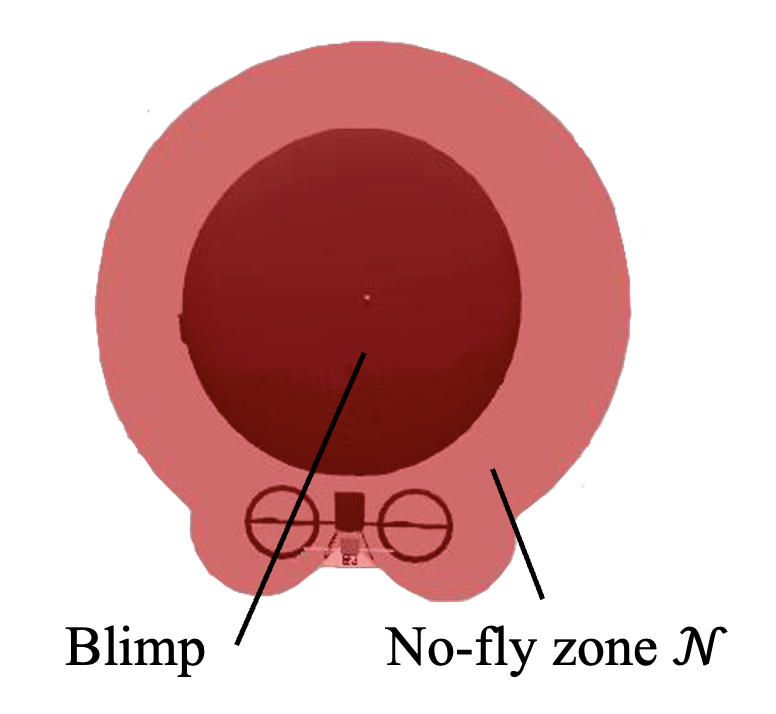}
        \caption{Cross-section view}
        \label{fig:no_fly_zone_cross}
    \end{subfigure}
    \hfill
    \begin{subfigure}[b]{0.25\textwidth}
        \centering
         \includegraphics[width=\linewidth]{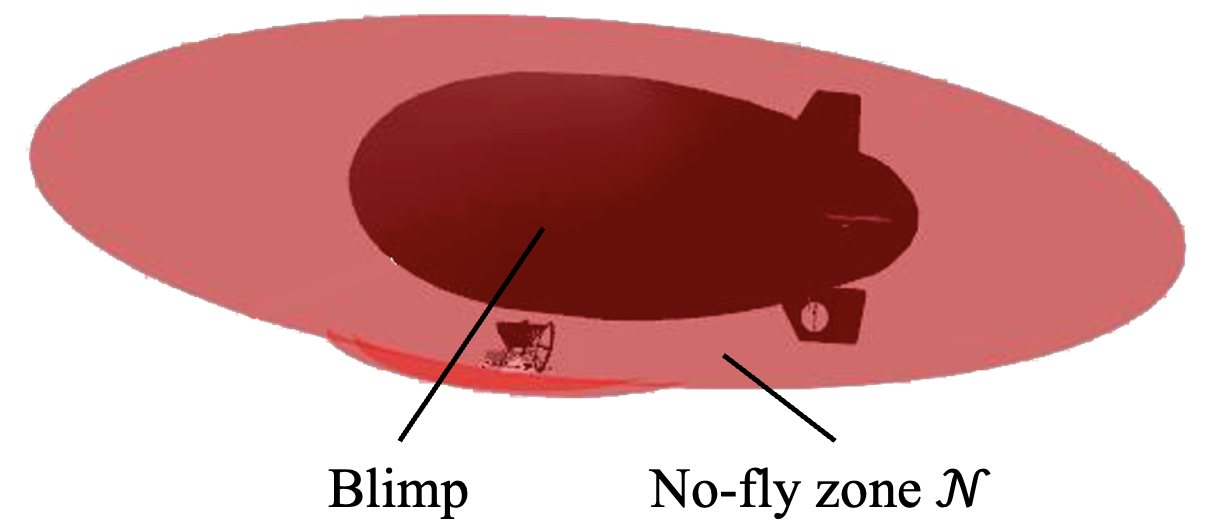}
        \caption{Side view}
    \label{fig:no_fly_zone_side}
    \end{subfigure}    
    \caption{No-fly zone around the blimp.}
    \label{fig:no_fly_zone}
\end{figure}

We begin with the assumption that given a blimp and multi-rotor UAV docking between these two vehicles shall be achieved within a reasonable amount of time, i.e. less than a minute. Blimps have the ability to hover in one spot. However, we consider docking under these circumstances as trivial and focus on a harder task, a moving blimp. In the scope of this work, we consider a scenario where the blimp travels on a straight line trajectory. This is a common assumption for blimps surveying systematically a predefined area. Nevertheless, our methods are designed in such a way that they should be able to handle arbitrary blimp motion. This depends on the versatility of the dataset used for the training of the TCN predicting the blimp trajectory.
During the docking procedure, wind gusts, described by a velocity vector $\mathbf{v}_{g}(t)$, can occur leading to its deviation from the original trajectory. They are modeled as isolated events, i.e. it is assumed that the effect of a wind gusts has fully subsided before the next gust. The objective of the multi-rotor UAV is to reach the docking port without penetrating a no-fly zone $\mathcal{N}$ encapsulating the blimp, see Figure \ref{fig:no_fly_zone}.

 The docking port is located in the gondola on the downside of the blimp. Docking is considered successful when the multi-rotor UAV reaches a point located $0.3\si{m}$ below the docking port with a lateral tolerance of $0.3\si{m}$ and a vertical tolerance of $0.1\si{m}$.  

\subsection{Overview of our Proposed Method}

\begin{figure}[htbp]
    \centering
    \includegraphics[width=0.5\textwidth]{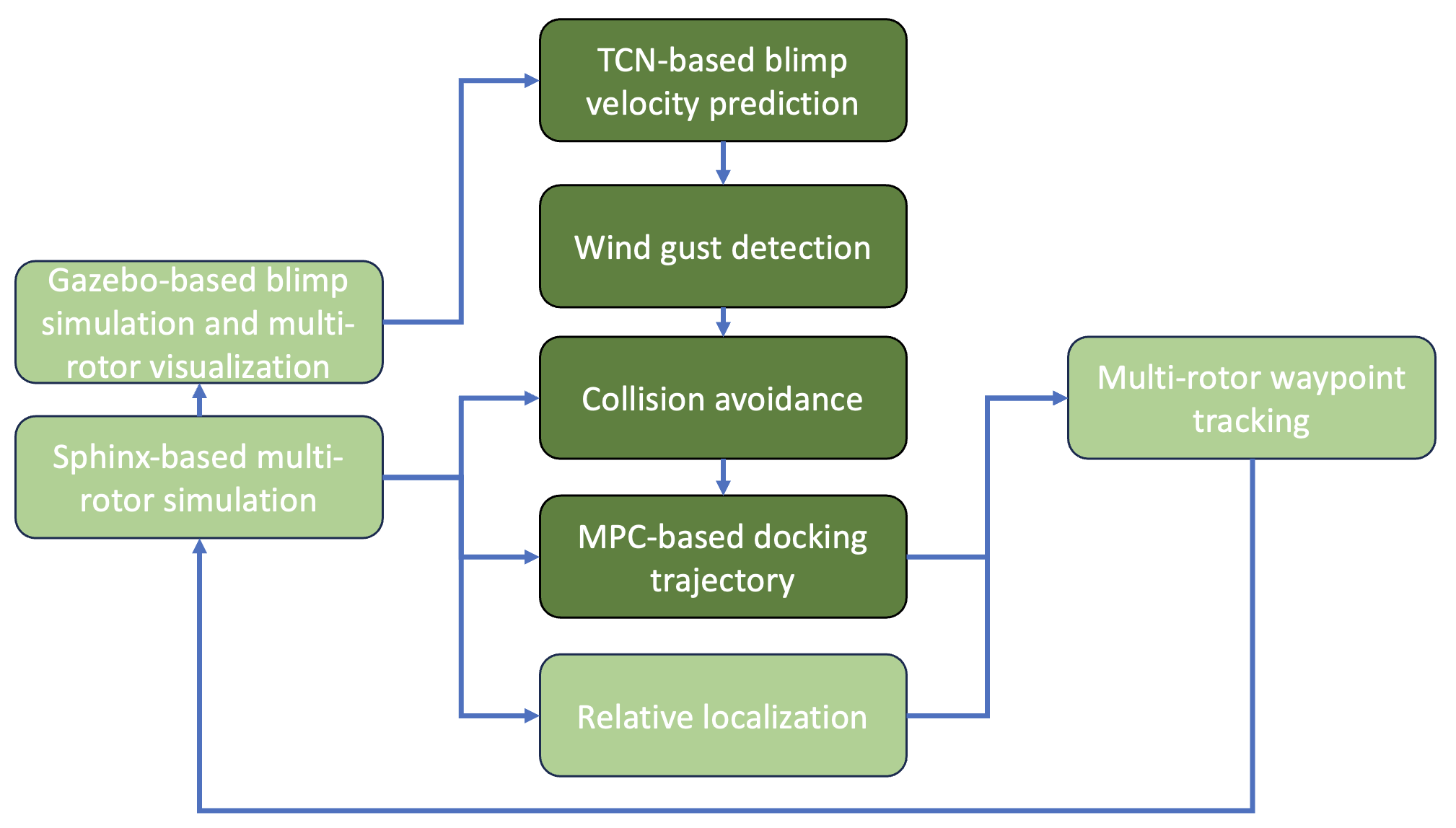}
    \caption{System overview, colored light green are components relying on existing work, colored dark green are components containing contributions of this work that are explained in closer detail.}
    \label{fig:system_overview}
\end{figure}

 The blimp's response to wind disturbances is predicted using a Temporal Convolutional Network (TCN), and the resulting predicted trajectory is incorporated into a Model Predictive Control (MPC) framework. This MPC computes a feasible, collision-free trajectory that guides the multi-rotor UAV to the docking port.
Collision avoidance is achieved using our novel CETH method. The novelty lies in the aspect that the method allows close range maneuvering of the multi-rotor UAV to the blimp only within a conical approach corridor guiding the UAV to the docking port. The TCN-based trajectory prediction is evaluated using a wind-gust detection algorithm, we introduced in this work. Upon detecting a gust, the approach corridor is dynamically deactivated to rapidly steer the multi-rotor UAV away from the blimp, thereby reducing the risk of collision.
Relative localization between the multi-rotor platform and the blimp is achieved using fiducial markers (AprilTags) affixed near the docking interface. 
Observations of these markers are fused with the blimp’s shared state estimate that comprises pose and velocity via an Extended Kalman Filter (EKF) to ensure accurate relative positioning.
All inter-system communication is facilitated through the Robot Operating System (ROS) middleware. A schematic overview of the proposed docking system is presented in Figure~\ref{fig:system_overview}, with detailed descriptions of its individual components provided in the subsequent subsections.

\subsubsection{Wind Gust Definition}

In this work, we adopt a wind gust model for the blimp inspired by the gust load design methodology used in manned aircraft, as described in \cite{easa_cs23_amendment_4}. The scalar gust velocity profile \( v_{g}(t) \) follows a standardized \( 1 - \cos \) shape, ramping smoothly from zero to a maximum value \( v_{g,{max}} \) and back to zero.

Temporally, the gust event is segmented into three distinct phases. The initial period, denoted \( t_0 \), precedes the onset of the gust and extends until time \( t_1 \), the start of the gust. The gust itself spans a duration \( T_g \), resulting in the interval \( [t_1, t_1 + T_g] \), followed by the post-gust period until \( t_2 \). The direction of the gust, represented by the unit vector \( \mathbf{d}_g \in \mathbb{R}^3 \), is sampled component-wise from a uniform distribution over the interval \([-1, 1]\) and then normalized, capturing the isotropic nature of wind disturbances in free air.

Within the gust interval, the scalar velocity profile is defined by:
\begin{equation}
\small
v_{g}(t) = \frac{v_{g,{max}}}{2} \left(1 - \cos\left(2\pi \cdot \frac{t - t_{1}}{T_{g}} \right)\right), \quad \text{for } t \in [t_{1}, t_{1} + T_g]
\end{equation}

The resulting gust velocity vector applied to the blimp is then given by:
\begin{equation}
\mathbf{v}_g(t) = v_{g}(t)\mathbf{d}_g
\end{equation}

\subsection{Blimp Gust Response Prediction using a Temporal Convolutional Network }
In order to model the dynamics of the blimp's response to a wind gust, we leverage a TCN. The TCN processes a timeseries of $n_{pred} = 98$ of 6-dimensional feature vectors sampled at a frequency of $10\si{hz}$. The feature vectors are composed of the blimps velocity vector $\textbf{v}_{blimp} \in \mathbb{R}^3[\si{m}]$  and Euler angles $\boldsymbol{\varphi} = [\phi,\theta,\psi] \in\mathbb{R}^3[\si{rad}]$ where $\phi,\theta,\psi$ denote the roll, pitch, yaw angles in the global coordinate frame. 
The network architecture comprises four residual blocks built using causal, dilated one-dimensional convolutions. Specifically, the dilation factors for the four blocks are $2^0 = 1$, $2^1 = 2$, $2^2 = 4$, and $2^3 = 8$. The first block uses a kernel size of 1, while the remaining blocks use a kernel size of 9. All blocks contain 32 convolutional filters. The first residual block maps the 6-dimensional input to a 32-dimensional hidden representation. Intermediate blocks maintain this dimensionality, and the final block reduces the output back to 6 dimensions. The output is a vector containing again the blimp velocities and Euler angle, just like the feature vectors, also being of the same length. Residual connections are applied throughout, with $1 \times 1$ convolutions used to match channel dimensions when necessary. Each convolutional layer is followed by layer normalization and a ReLU activation function. All network weights were initialized using He normal initialization.
We employ recursive forecasting to extend the prediction horizon of the blimp’s dynamic response. This approach is essential due to the presence of slow dynamic modes in the blimp's behavior, particularly in response to wind disturbances, which necessitate long-term predictive capabilities. In recursive forecasting, the output of a single evaluation of the TCN is recursively fed back as input for subsequent predictions, enabling multi-step forecasting. To prevent discontinuities at the transition points between successive prediction segments, we implement a strategy that connects the value of the preceding forecast with an appropriate point in the subsequent prediction window. For this purpose smoothing techniques and a kink-free fade-in are applied. Implementation details can also be seen in the code provided with this work.

\subsubsection{Wind Gust Detection}
We analyze the predicted blimp velocities \( v_{pred,d}(k) \), where \( d \in \{x, y, z\} \) and \( k = 0, \ldots, n_{pred}-1 \), obtained from the temporal convolutional network (TCN), to detect wind gust events. As an initial step, we compute the mean predicted velocity in each spatial direction:

\begin{equation}
\bar{v}_{pred,d} = \frac{1}{n_{pred} } \sum_{k=0}^{n_{pred}-1} v_{\text{pred},d}(k)
\end{equation}

To enhance temporal localization, we implement a windowing strategy by partitioning the sequence of predicted velocities into \( W_g \in \mathbb{N} \) non-overlapping segments \( w_i \), each of length \( n_w \in \mathbb{N} \). Within each window \( w_i \), we compute the maximum absolute deviation of the predicted velocity from its corresponding mean value:

\begin{equation}
\Delta_d^{(i)} = \max_{t \in w_i} \left| v_{pred,d}(t) - \bar{v}_{pred,d} \right|
\end{equation}

A wind gust event is flagged if the maximum deviation \( \Delta_d^{(i)} \) in any direction \( d \in \{x, y, z\} \) exceeds a predefined threshold \( \delta_v \). In combination with the   window length $n_w$ this parameter can be set by the user to empirically adapt the detection sensitivity. Shorter and thus more windows allow a more accurate determination of  when the wind gust effects will have subsided. However, since in this case the mean  $\bar{v}_{pred,d}$  is  calculated over a shorter time period, deviations are typically small  requiring careful tuning of \( \delta_v \). To reduce the likelihood of false positives caused by transient oscillations in the TCN's recursive prediction process, especially particularly near the boundary of the forecast horizon, the temporal range for gust detection is intentionally constrained by a value $t_{max, g}$.

\subsection{Collision-free Trajectory Planning using a Model Predictive Controller (MPC)}\label{sec:trajectory_mpc}
To generate collision-free and constraint-aware trajectories for the  multi-rotor UAV, we employ a discrete-time linear Model Predictive Control (MPC) framework. The controller solves a constrained quadratic optimization problem using the OSQP solver. This formulation yields a sequence of control inputs that track a reference trajectory of the system states defined for the MPC.

The state vector of the system at time step \( k \) is defined as \( \mathbf{x}_k = [\mathbf{p}_k^\top \mathbf{v}_k^\top]^\top \in \mathbb{R}^6 \), where \( \mathbf{p}_k \in \mathbb{R}^3 \) and \( \mathbf{v}_k \in \mathbb{R}^3 \) represent the position and velocity of the multi-rotor UAV, respectively. The control input vector is given by \( \mathbf{u}_k = [\mathbf{a}_k^\top \mathbf{f}_k^\top]^\top \in \mathbb{R}^6 \), where \( \mathbf{a}_k \in \mathbb{R}^3 \) are the free optimization variables corresponding to commanded accelerations, and \( \mathbf{f}_k \in \mathbb{R}^3 \) are fixed external forces. The latter are used as part of our novel CETH method to model the tangential hull, as detailed in Section \ref{sec:tangential_band}.

The system dynamics are modelled using a discretized linear time-invariant system of the form:
\[
\mathbf{x}_{k+1} = \mathbf{A}_d \mathbf{x}_k + \mathbf{B}_d \mathbf{u}_k,
\]
with the discrete-time system matrices defined as:
\[
\mathbf{A}_d = \begin{bmatrix}
\mathbf{I}_3 & \Delta t \cdot \mathbf{I}_3 \\
\mathbf{0}_3 & \mathbf{I}_3
\end{bmatrix}, \quad
\mathbf{B}_d = \begin{bmatrix}
\frac{1}{2} \Delta t^2 \cdot \mathbf{I}_3 & \frac{1}{2} \Delta t^2 \cdot \mathbf{I}_3 \\
\Delta t \cdot \mathbf{I}_3 & \Delta t \cdot \mathbf{I}_3
\end{bmatrix},
\]
where \( \Delta t \) denotes the MPC discretization timestep.

To ensure the fixed nature of the external force component \( \mathbf{f}_k \), appropriate box constraints are imposed on the control inputs. Specifically, the minimum and maximum bounds on the control input are defined as:
\[
\mathbf{u}_{\min,k} = \begin{bmatrix} \mathbf{a}_{\min} \\ \mathbf{f}_k \end{bmatrix}, \quad  
\mathbf{u}_{\max,k} = \begin{bmatrix} \mathbf{a}_{\max} \\ \mathbf{f}_k \end{bmatrix},
\]
thereby constraining \( \mathbf{f}_k \) to a constant value within the optimization problem.

The MPC problem is then formulated as the following finite-horizon quadratic program:
\begin{equation}
\begin{split}
\begin{array}{ll}
\{\mathbf{x}^*_k, \mathbf{u}^*_k\}_{k=0}^N = \underset{\{\mathbf{u}_k\}_{k=0}^{N-1}}{\text{arg min}} & J\\
\text{subject to} \quad & \mathbf{x}_{k+1} = \mathbf{A}_d \mathbf{x}_k + \mathbf{B}_d \mathbf{u}_k, \\
& \mathbf{x}_{\min} \le \mathbf{x}_k \le \mathbf{x}_{\max}, \\
& \mathbf{u}_{\min} \le \mathbf{u}_k \le \mathbf{u}_{\max}, \\
& \mathbf{x}_0 = \bar{\mathbf{x}}, \\
& \mathbf{z}_0 = \bar{\mathbf{z}},
\end{array}
\end{split}
\label{eq:MPC}
\end{equation}
with 
\begin{equation}
\small
 J = (\mathbf{x}_N - \mathbf{z}_N)^T \mathbf{Q}_N (\mathbf{x}_N - \mathbf{z}_N) +  \sum_{k=0}^{N-1} (\mathbf{x}_k - \mathbf{z}_k)^T \mathbf{Q} (\mathbf{x}_k - \mathbf{z}_k)  + \mathbf{u}_k^T \mathbf{R} \mathbf{u}_k 
 \label{eq:cost_function}
\end{equation}
In this formulation, $\mathbf{z}_k =  [\mathbf{p}_{blimp,k}^\top, \mathbf{v}_{blimp,k}^\top]^\top \in \mathbb{R}^6 $ denotes the reference trajectory, where the $\mathbf{p}_{blimp}$ and $\mathbf{v}_{blimp}$ denote the blimp's position and velocity as predicted by the TCN.   $\mathbf{p}_{blimp}$ is obtained from $\mathbf{v}_{blimp}$ via the Simpson integration method. 
We deploy two sources for $\mathbf{v}_{blimp}$. Following a constant-velocity based approach, one option is $\mathbf{v}_{blimp,k} = \mathbf{v}_{blimp,0},k= 0\ldots,N $ with $\mathbf{v}_{blimp,0}$ being the blimp's velocity at the beginning of the optimization cycle. Constant velocity models are one of the widely used models in target tracking \cite{li2003}. The second option is to use the TCN prediction of the blimps velocity, so $\mathbf{v}_{blimp,k} = \mathbf{v}_{pred,k},k= 0\ldots,N $ which allows for a more accurate prediction of the blimp's velocity and thus to better performance as is shown in Section \ref{sec:experiments}. The matrices \( \mathbf{Q} \), \( \mathbf{Q}_N \), and \( \mathbf{R} \) are positive semi-definite weighting matrices that penalize deviations from the reference trajectory and excessive control effort. Additionally, whenever the multi-rotor enters the tangential hull used for collision avoidance (see Section \ref{sec:tangential_band} for further details), the weighting matrices \( \mathbf{Q} \) and \( \mathbf{Q}_N \) are adaptively reduced to \( \mathbf{Q}_{\min} \) and \( \mathbf{Q}_{N,\min} \), respectively.

\subsection{Collision Avoidance} \label{sec:tangential_band}
\begin{figure}[htbp]
    \centering
    \includegraphics[width=0.45\textwidth]{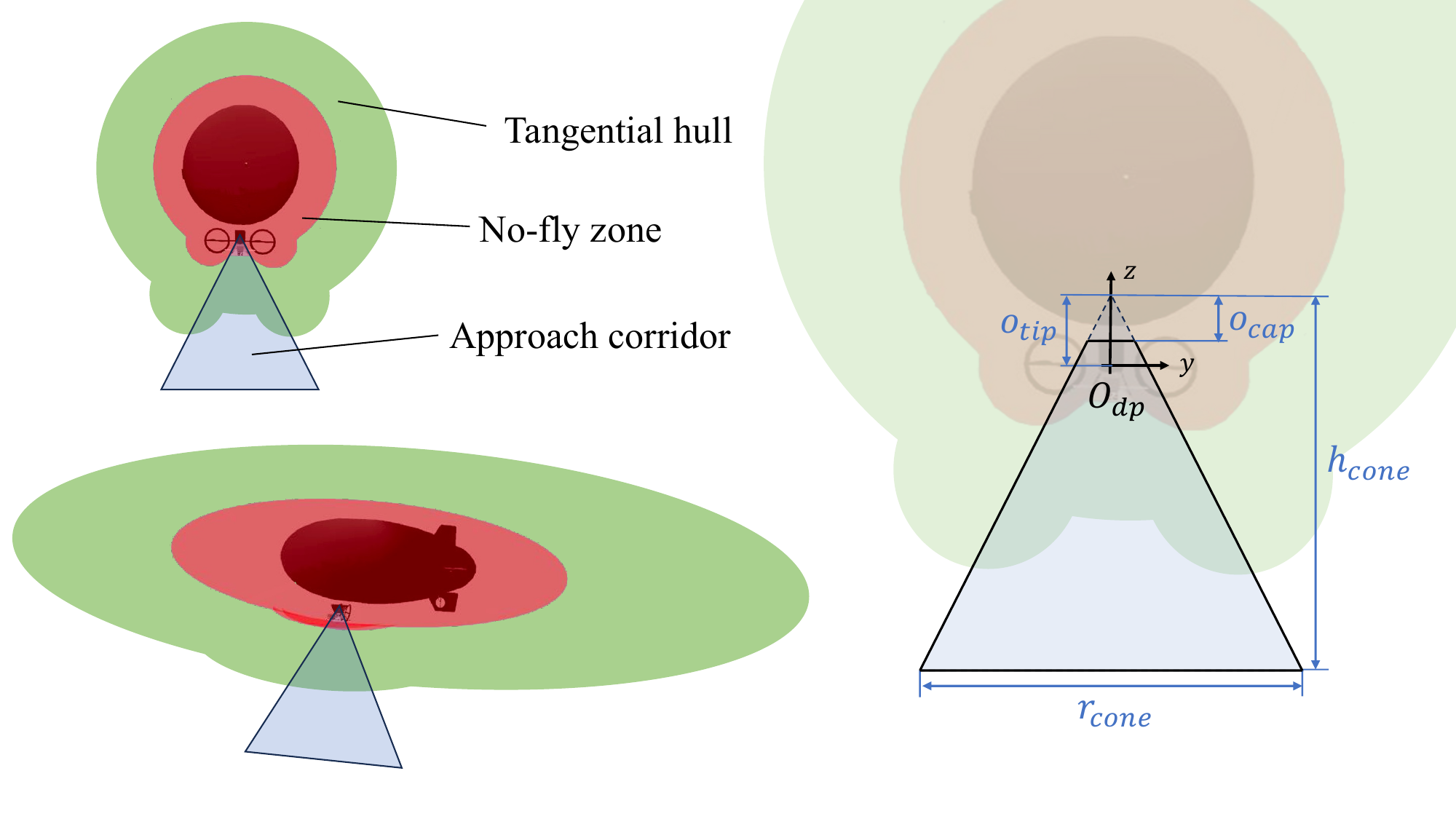}
    \caption{Illustration of the corridor-enhanced tangential hull method.}
    \label{fig:ceth}
\end{figure}

 We propose a novel technique called the \textit{Corridor-enhanced Tangential Hull} (CETH) method for obstacle avoidance.
To this end, we build on the tangential band method presented in \cite{tallamraju2018}. Here, a 2D obstacle is enclosed in a tangential band exerting additional forces on the approaching UAV. There are two components, a repulsive force pushing the UAV away and a force tangential to the obstacle surface in order to drive the UAV around the obstacle, like it is transported on a band, hence the name. In our method,  we consider  3D obstacles such as the no-fly zone $\mathcal{N}$ that are enclosed by a tangential hull. We  introduce the following novelties. i) We integrate an approach corridor penetrating the tangential hull to open a passageway for the multi-rotor UAV towards the docking port.  
Geometrically, the approach corridor is modelled as a truncated cone with a base radius \( r_{\text{cone}} \) and height \( h_{\text{cone}} \), with its apex located slightly above the docking port of the blimp in a distance $o_{\text{tip}}$ and its central axis pointing downwards, parallel to the blimp body's vertical axis. The truncated part of the cone is measured from the tip and has a height of $h_{cap}$. It is illustrated by Figure \ref{fig:ceth}. Depending on the obstacle avoidance settings, the approach corridor is closed whenever a wind gust is detected thus ensuring reliable collision avoidance. ii) We design the tangential forces in such a way that they guide the multi-rotor UAV towards the docking port and thus inside the approach corridor.
 iii) a discretization of the 3D space around the no-fly zone for which variables required for the collision avoidance method are precomputed. This replaces time-consuming online-computation required for 3D obstacles with a quick and simple table look-up. Overall, this allows faster execution time of the trajectory MPC approach presented in Section \ref{sec:trajectory_mpc}. 
 
\subsubsection{Repulsive Force Model}

Let \( d \in \mathbb{R}_{\geq 0} \) denote the Euclidean distance from the multi-rotor UAV to the surface of the no-fly zone $\mathcal{N}$. The repulsive force \( \mathbf{F}_{\mathrm{rep}} \in \mathbb{R}^3 \) acting on the multi-rotor is defined as follows.

\begin{equation}
\mathbf{F}_{\mathrm{rep}} = 
\begin{cases}
\min\left( \phi(d), F_{\text{max,rep}} \right)  \mathbf{n}_{\mathrm{rep}}, & \text{if } d < d_b \\
0, & \text{otherwise}
\end{cases}
\end{equation}
where,  \( \mathbf{n}_{\mathrm{rep}} \) is the unit vector pointing from the surface of the no-fly zone $\mathcal{N}$ towards the multi-rotor UAV, \( d_b \) is the repulsion cutoff distance, \( F_{\text{max},rep} \) is the maximum repulsive force magnitude, \( \phi(d) \) is a distance-dependent potential function.

The potential magnitude \( \phi(d) \) is given by:
\begin{equation}
\phi(d) = 
\begin{cases}
k_{\mathrm{rep}}  U(d), & \text{if } d < d_b \\
0, & \text{otherwise}
\end{cases}
\end{equation}
where \( k_{\mathrm{rep}} > 0 \) is a repulsion gain, and \( U(d) \) is a cotangent-based shaping function that increases steeply as \( d \) approaches a minimum safety distance \( d_{\min} \):
\begin{equation}
U(d) = \frac{\pi}{2(d_b - d_{\min})} \left[ \cot(z(d)) + z(d) - \frac{\pi}{2} \right]
\end{equation}

The intermediate variable \( z(d) \) is defined piecewise as:
\begin{equation}
z(d) = 
\begin{cases}
\frac{\pi}{2}  \frac{0.1}{d_b - d_{\min}}, & \text{if } d < d_{\min} \\
\frac{\pi}{2}  \frac{d - d_{\min}}{d_b - d_{\min}}, & \text{if } d_{\min} \leq d < d_b
\end{cases}
\end{equation}

This formulation ensures that the repulsive potential grows sharply and smoothly as the multi-rotor UAV nears the no-fly zone $\mathcal{N}$.

\subsubsection{Tangential Force Model}
 The tangential direction is derived by projecting the target vector, i.e. the vector from the multi-rotor's position to the docking port onto the no-fly zone's local tangent plane.

Let \( \mathbf{p} \in \mathbb{R}^3 \) be the multi-rotor's current position, \( \mathbf{p}_{\text{dp}} \in \mathbb{R}^3 \) the position of the docking port, and \( \mathbf{n} \in \mathbb{R}^3 \) the unit surface normal of the no-fly zone $\mathcal{N}$ at \( \mathbf{p} \). The vector from the multi-rotor to the docking port is
\begin{equation}  
\mathbf{v} = \mathbf{p}_{\text{dp}} - \mathbf{p}.
\end{equation}

We project \( \mathbf{v} \) onto the tangent plane orthogonal to \( \mathbf{n} \):
\begin{equation}  
\mathbf{v}_{\perp} = \mathbf{v} - (\mathbf{v} \cdot \mathbf{n}) \mathbf{n}.
\end{equation}

The tangential direction \( \hat{\mathbf{t}} \) is then given by
\begin{equation}  
\hat{\mathbf{t}} = 
\begin{cases}
\dfrac{\mathbf{v}_{\perp}}{\|\mathbf{v}_{\perp}\|}, & \text{if } \|\mathbf{v}_{\perp}\| > \epsilon, \\
\text{fallback direction}, & \text{otherwise}.
\end{cases}
\end{equation}
Here, $\epsilon$ denotes a small value greater than zero. This is necessary to handle the edge case where there is no component tangential to the no-fly zone's surface. If this edge case occurs,  a fallback direction is used which is either the previous tangential direction or just the direction to the docking port to initiate motion of the multi-rotor UAV in the hope to switch to a position where the tangential direction can be calculated. 
The tangential force \( \mathbf{F}_{\text{tang}} \) is applied only when the multi-rotor is within a distance band \( d \leq d_{\text{band}} \) from the no-fly zone surface:
\begin{equation}  
\mathbf{F}_{\text{tang}} = 
\begin{cases}
\min\left(k_{\text{max,tang}} \cdot \|\nabla J\|, F_{\text{max}}\right) \cdot \hat{\mathbf{t}}, & \text{if } d \leq d_{\text{band}}, \\
\mathbf{0}, & \text{otherwise},
\end{cases}
\end{equation}
where \( k_{\text{tang}} \) is a gain coefficient, \( \nabla J \) represents the summed gradient of the cost function $\nabla J = \frac{\partial J}{\partial x(N)} + \frac{\partial J}{\partial u(N-1)}$ used in \eqref{eq:cost_function}, and \( F_{\text{max,tang}} \) is the maximum allowable tangential force. 
The magnitude of the cost function's gradient $\Vert\nabla J\Vert$ is included in the tangential force in order to achieve a more pronounced effect if the multi-rotor UAV is far away from the docking port than in later stages of the docking procedure when the multi-rotor UAV has moved closer to the docking port.

\subsubsection{Computation of Forces via Look-up Table}
For the computation of the repulsive and tangential forces, the distance of the multi-rotor UAV to the no-fly zone $\mathcal{N}$ is required. However, since the shape of the no-fly zone is non-convex, online computation is too time consuming since iterative methods need to be employed. For this reason, we discretize a portion of the 3D space containing the no-fly zone in order to obtain a grid of 3D cells. We then compute the required variables such as distance to the obstacle surface, the closest point on the obstacle surface or the direction of the repulsive force in advance for the each grid point and store the result in a hashable data structure. During runtime we then map the position of the multi-rotor UAV to a grid point and look up the required variables required for the computation of the forces of the CETH method.

\subsubsection{Safety Position} \label{sec:safety_pos}
In the event of a wind gust, the optional possibility is given to fly the multi-rotor UAV to a safety position. This safety position is calculated by means of a sliding target point which slides along the predicted trajectory of the blimp until safety position $p_{safe}$ is reached. The sliding motion only occurs during the presence of a wind gust. When the wind gust is over, the target point is reset. Details can be explored in the code provided online (see link on first page).

\subsection{EKF-Based Docking Port Localization}
To estimate the docking port position with high accuracy, we employ an Extended Kalman Filter (EKF) that fuses the  GPS-based state estimate shared by docking port mounted on the blimp with accurate, but intermittent, visual measurements from AprilTag detections taken by the multi-rotor UAV's camera. The EKF estimates a correction vector $\mathbf{x}_{k,bias} \in \mathbb{R}^3$ that accounts for the offset between the position estimate of the docking port and the estimated position of the multi-rotor's camera.

A low-pass filter is applied to smooth the output.

\section{Implementation}\label{sec:implementation}
\subsection{General}
We design our experiments to demonstrate the following. 
\begin{itemize}
\item \textit{TCN performance}: Our recursive forecasting method allows for sufficiently accurate predictions of the TCN for multiple forecasting steps compared to a constant velocity model.
\item \textit{CETH performance}: The efficacy of the CETH method we developed for obstacle avoidance while allowing close-range maneuvering of the multi-rotor UAV in the approach corridor. 
\item \textit{TCN benefit}: The benefit of using our TCN-based velocity prediction to calculate the blimp trajectory over a constant velocity model.

\end{itemize}
\subsection{Training and Simulation Hardware}
All trainings are run on a desktop computer with Ubuntu 20, AMD Ryzen threadripper 3960x 24-core processor, 128 GB RAM, 6TB SSD. This computer is also used for the simulation experiments in combination with a second desktop computer with the following specifications. Ubuntu 20, AMD Ryzen Threadripper 3960X 24-Core CPU, NVIDIA GeForce RTX 2080 Ti GPU, 128GB RAM and 2TB SSD. The first computer hosts the Sphinx simulator for the simulation of the Anafi drone as well as the majority of our framework's other components. On the second computer the airship simulation \cite{price2022} is run with all relative localization functionalities. This distributed setup was necessary in order to be able to achieve a real time factor of 1 for both simulators, Sphinx and Gazebo.
\begin{figure*}[]
    \centering
    \includegraphics[width=0.8\textwidth]{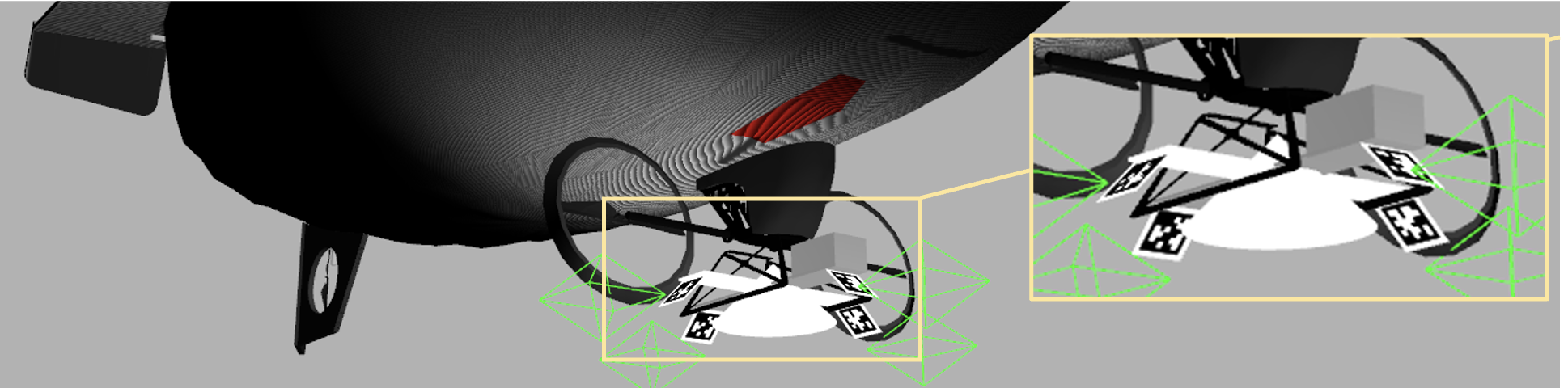}
    \caption{Design of the docking port in simulation. AprilTags are used for relative localization.}
    \label{fig:docking_port_design_sim}
\end{figure*}
\subsection{Simulation Frameworks}
We implemented our algorithms  using Python3 in the ROS noetic middleware \cite{quigley2009}.  For the simulation of the blimp we used the airship simulation \cite{price2022} which is realized in the physics simulator Gazebo 11 that is integrated in ROS noetic. For the simulation of the multi-rotor UAV we relied on the mirroring technique presented in \cite{goldschmid2025} in order to include a mirrored instance of the commercial Anafi drone simulated in the partially proprietary Sphinx simulator into the Gazebo-based airship simulation.  The Anafi drone is also used in our real world experiments.

\subsection{Docking Port Design}
We envision a docking port mounted at the gondola of the blimp between the main lateral thrusters. AprilTags are attached to a carrier structure on angled plates to provide good visibility from all possible approach angles as illustrated by Figure \ref{fig:docking_port_design_sim}. The central locking device which physically connects the multi-rotor vehicle with the airhsip is expected to sit in the middle of the structure carrying the AprilTags.

\subsection{Dataset of Blimp Responses to Wind Gusts}\label{sec:blimp_datasets}
We recorded numerous wind gusts and their effect on the blimp in simulation in multiple separate episodes. On the one hand, these episodes are used for the creation of the TCN's training dataset and also used for its evaluation. Common to all  recorded episodes designated for these tasks is that the wind gust direction is sampled randomly from a uniform distribution. However, the maximum gust velocity $v_{g,max}$ varies. These episodes were decomposed into $147840$ samples constituting the training dataset. On the other hand, longer episodes were recorded for the evaluation of the trajectory planning MPC. Their directions are sampled approximately equidistantly on a unit sphere and only a wind gust velocity $v_{g,max}$ is considered. Lower wind gust velocities are not examined since they constitute a simpler problem and are thus not used for verification. Details are summarized in Table \ref{tab:dataset}.

\begin{table}[h!]
\centering
\small
\caption{Parameters of wind gusts included in the dataset used for the training of the TCN and for evaluation.}
\label{tab:dataset}
\begin{tabular}{l c c c c c}
\toprule
Dataset & $v_{g,{max}} [\si{m/s}]~~$ & No. in dataset & $~~t_0[\si{s}]~~$ & $~~T_g [\si{s}]~~$ & $~~t_2 [\si{s}]~~$ \\
\midrule
\multirow{4}{*}{Training}
& 1 & 695 & 10 & 4 & 46 \\
& 2 & 186 & 10 & 4 & 46 \\
& 3 & 101 & 10 & 4 & 46 \\
& 4 & 900 & 10 & 4 & 46 \\
\midrule
Evaluation & 4 & 10 & 10 & 4 & 196 \\
\bottomrule
\end{tabular}
\end{table}
\subsection{Training of the TCN}
The TCN  was trained using the Adam optimizer with a learning rate of 0.0005, decaying by a factor of 0.1 at epochs 12, 24, and 30. A batch size of 64 was used, and gradient clipping with a maximum norm of 0.8 was applied to ensure stable optimization. The training objective was an ordered multi-output regression loss function.
To ensure numerical stability and promote convergence, all input features and target variables were normalized to the $[0, 1]$ range using min-max scaling based on the empirical ranges observed in the training dataset. 

\subsection{MPC-based Trajectory Planning of the Multi-Rotor}
The parameters chosen for the MPC  are summarized in Table \ref{tab:mpc_params}. 
\begin{table}[h!]
\centering
\small
\caption{MPC parameters used in the trajectory planning algorithm. Vector quantities are element-wise.}
\label{tab:mpc_params}
\begin{tabular}{l c c}
\toprule
Parameter & Value & Unit \\
\midrule
Time step $\Delta t$ & 0.1 & \si{s} \\
Prediction horizon $N$ & 15 & -- \\
\midrule
Minimum acceleration $\mathbf{a}_{min}$ & $[-5,\,-5,\,-5]$ & \si{m/s^2} \\
Maximum acceleration $\mathbf{a}_{max}$ & $[5,\,5,\,5]$ & \si{m/s^2} \\
Minimum velocity $\mathbf{v}_{min}$ & $[-8,\,-8,\,-4]$ & \si{m/s} \\
Maximum velocity $\mathbf{v}_{max}$ & $[8,\,8,\,4]$ & \si{m/s} \\
\midrule
State cost weights $\mathbf{Q}$ & $\mathrm{diag}\left(\mathbf{0}_6\right)$ & -- \\
Terminal cost weights $\mathbf{Q}_N$ & $\mathrm{diag}\left(100 \cdot \mathbf{1}_3,\; 10 \cdot \mathbf{1}_3\right)$ & -- \\
Control cost weight $\mathbf{R}$ & $\mathrm{diag}\left(1 \cdot \mathbf{1}_3,\mathbf{0}_3\right)$ & -- \\
Scale factor $\alpha_{QN}$ & 0.002 & -- \\
\bottomrule
\end{tabular}
\end{table}
The parameters chosen for the CETH method used for collision avoidance are presented in Table \ref{tab:tangential_band_params}. The parameters of our wind gust detection approach are listed in Table \ref{tab:gust_detection_params}.

\begin{table}[h!]
\centering
\small
\caption{Parameters used for the CETH method for collision avoidance.}
\label{tab:tangential_band_params}
\begin{tabular}{l c c}
\toprule
Parameter & Value & Unit \\
\midrule
Repulsive gain $k_{\text{rep}}$ & 1 & -- \\
Tangential gain $k_{\text{tang}}$ & 10 & -- \\
Tangential band width $d_{\text{band,tang}}$ & 5 & \si{m} \\
Repulsive band width $d_{\text{band,rep}}$ & 5.5 & \si{m} \\
Minimum distance $d_{\text{min}}$ & 1 & \si{m} \\
Max repulsive force $F_{\text{max},\text{rep}}$ & 6 & \si{N} \\
Max tangential force $F_{\text{max},\text{tang}}$ & 2 & \si{N} \\
Cone radius $r_{\text{cone}}$ & 8 & \si{m} \\
Cone height $h_{\text{cone}}$ & 7 & \si{m} \\
Cone tip offset $o_{\text{tip}}$ & 0.12 & \si{m} \\
Cone cap height $h_{\text{cap}}$ & 0.15 & \si{m} \\
\bottomrule
\end{tabular}
\end{table}

\begin{table}[h!]
\centering
\small
\caption{Parameters for detecting wind gusts.}
\label{tab:gust_detection_params}
\begin{tabular}{l c c}
\toprule
Parameter & Value & Unit \\
\midrule
Number of windows for gust detection $W_g$ & 10 & -\\
Temporal gust detection boundary $t_{max, g}$ & 6 & \si{s} \\
Gust detection threshold  $\Delta_d$ & 0.9 & \si{m/s} \\
\bottomrule
\end{tabular}
\end{table}

\subsection{Evaluation of the TCN}
We assess the performance of the TCN employed as a velocity prediction model for the blimp  in comparison to a  constant velocity model. We choose this comparison as constant velocity models are one of the commonly used models in target tracking \cite{li2003}. They are simple and computationally efficient. We compute the rolling mean squared error (RoMSE) over a time frame of $\Delta t N$, where $\Delta t$ denotes the time step of the model predictive controller, and $N$ is the number of future prediction steps.

The predictions generated by both the constant velocity model and the TCN are evaluated against the ground-truth velocity data recorded over $T$ time steps in the simulation. For each trajectory, the first $K$ time steps are omitted, as they are required as input to the TCN. Consequently, the number of prediction windows available is given by $W = T - N - K$.

The rolling MSE for a given direction of motion $d \in \{x, y, z\}$ is then defined as:
\begin{equation}
\text{Rolling MSE}_d = \frac{1}{W \cdot N} \sum_{i=K}^{T - N - 1} \sum_{j=0}^{N - 1} \left\| \hat{v}_{d, i + j} - v_{d, i + j} \right\|^2,
\label{eq:rolling_mse}
\end{equation}
where $\hat{v}_{d}$ denotes the predicted velocity in direction $d$ (from either the constant model or the TCN), and $v_{d}$ denotes the corresponding ground truth.

To evaluate overall performance, we compute the rolling MSE for a set of simulated wind gust conditions and report the mean error across all such scenarios.

\subsection{Simulation Experiments} \label{sec:simulation_experiments}
We introduce three different categories of experiments for testing the CETH performance as well as the benefit of using the TCN-based velocity model over the constant velocity model. 
\begin{table}[h!]
\centering
\small
\caption{Experiment categories and scenarios.}
\label{tab:sim_exps_overview}
\begin{tabular}{p{2.5cm} p{2.5cm} p{3.2cm}}
\toprule
Category & Subcategories & Description \\
\midrule
CETH inactive & \begin{tabular}{@{}l@{}}No abort\\ Safety position\end{tabular} & \begin{tabular}{@{}l@{}}CETH method disabled,\\ illustrates need for \\collision avoidance.\end{tabular} \\
\midrule
\begin{tabular}{@{}l@{}}CETH active\\ const. vel.\end{tabular} & \begin{tabular}{@{}l@{}}No abort\\ Abort\\ Safety position\end{tabular} & \begin{tabular}{@{}l@{}}CETH method active, \\constant velocity model \\ for blimp trajectory,\\ demonstrating efficacy \\ of the CETH method\end{tabular} \\
\midrule
\begin{tabular}{@{}l@{}}CETH active\\ TCN vel.\end{tabular} & \begin{tabular}{@{}l@{}}No abort\\ Abort\\ Safety position\end{tabular} & \begin{tabular}{@{}l@{}}CETH method active, \\TCN velocity model \\ for blimp trajectory,\\ showing benefits \\ of the TCN vel. over \\ the const. vel. model.\end{tabular} \\
\bottomrule
\end{tabular}
\end{table}
Each category has several subcategories, \textit{No abort}, \textit{Abort} and \textit{Safety position}, as summarized in Table \ref{tab:sim_exps_overview}. In \textit{No abort}, the docking procedure is continued also in the event of  a wind gust. In \textit{Abort}, the approach corridor is closed for experiments involving the CETH method. In the subcategory \textit{Safety position}, the multi-rotor UAV travels to a safety position as described in Section \ref{sec:safety_pos}. Each subcategory is tested with pre-recorded blimp trajectories included in the evaluation dataset presented in Section \ref{sec:blimp_datasets}. In each scenario the initial position of the blimp is $[0\si{m},0\si{m},30\si{m}]^T$ and the one of the multi-rotor UAV is at $[50\si{m},50\si{m},70\si{m}]^T$. The initial positions are chosen in such a way that the multi-rotor UAV meets the blimp while the wind gust has strongest effect on the blimp. Furthermore, the shortest trajectory towards the docking port leads through the no-fly zone $\mathcal{N}$. For this reason, it can be considered as a scenario with guaranteed collision if no collision avoidance method is deployed making it ideal for testing the performance of the CETH method.

Criteria to evaluate the docking performance are the percentage of trials ending in successful docking as described in Section \ref{sec:problem_statement} without occurrence of a collision. Furthermore, we determine the mean and standard deviation as well as the minimal and maximal values for the duration required until the docking criteria are met.

During the conduction of the simulation experiments, initially the first position element $\mathbf{p}^{*}_1$ from $\mathbf{x}^{*}_1 = [\mathbf{p}^*_1 \mathbf{v}^*_1]$ of the optimal trajectory computed by the trajectory planning MPC was sent as target position $\mathbf{p}_{t}$ to the waypoint controller of the Anafi drone. However, with this setting accurate tracking of the docking port was not possible. This is due to the tracking error which is present for moving target waypoints as outlined in \cite{goldschmid2025}. To resolve this issue, a weighted average between $\mathbf{p}^*_6$ and $\mathbf{p}^*_7$ is published whenever the multi-rotor UAV is inside the approach corridor. If it is inside the tangential hull, it is $\mathbf{p}_t =\mathbf{p}_{uav} +  6.5 \|\mathbf{p}^*_N -\mathbf{p}^*_0\| (\mathbf{p}^*_N -\mathbf{p}^*_0)$, a value which has been found empirically.

\subsection{Hardware for Real World Experiments}
For the real world experiments, two laptops were used hosting the framework in a distributed fashion. The first laptop hosted all required components of our framework whereas the second laptop was included to handle the controllers and communication of the multi-rotor UAV.  The first laptop has the following specifications. Intel Core i7 CPU, Nvidia T500 GPU, 64GB RAM, 2TB SSD. The second laptop features the following specifications. Intel Core i7 CPU, HD Intel Graphics 520, 32GB RAM, 500GB SSD. The drone used was the commercial drone Anafi by Parrot Drone SAS. 

\subsection{Real World Experiments}
Since ground truth data of wind gusts is typically difficult to obtain, we chose the following approach for our real world experiments. We selected one of the  blimp trajectories that was recorded in simulation and included in the training data set introduced in Section \ref{sec:blimp_datasets}. Besides the state information of the blimp, this recorded episodes also contain ground truth information of the wind gust, i.e. direction and velocity. This trajectory was then replayed during the real-world experiment. It  acts as a virtual blimp for our framework controlling the  Anafi drone's docking approach while at the same time providing information about the wind gust. 
We argue that the virtual blimp is a suitable replacement for a real blimp since the fidelity of the blimp simulation we used to record the blimp's trajectory  was extensively validated in real-world experiments in \cite{price2022}. For the real world experiments, the parameter cone tip offset was set to $o_{tip}=0.2$. 

In simulation, two MPCs were used. First, our trajectory planning MPC and then the low-level MPC presented in \cite{goldschmid2025} to achieve accurate waypoint tracking of the simulated Anafi drone. 
However, in the real-world experiments, we did not use the low-level MPC for waypoint following of the physical Anafi drone. The reason was that manual override of control commands by a human pilot was not implemented which we considered to be too unsafe for our experiments.  Instead, we relied on a PID-based controller providing this functionality for the tracking of the waypoints generated by our trajectory planning MPC. 
However, as outlined in Section  \ref{sec:simulation_experiments} the target position $\mathbf{p}_t$ is calculated from heuristics such as a weighted average between two consecutive positions in the calculated MPC trajectory.  In order to achieve a similar tracking behaviour of the PID controller, these heuristics had to be slightly adapted as can be seen in the code provided with this work.

\section{Results}\label{sec:experiments}
\subsection{Evaluation of the TCN and the Recursive Forecasting Method in Simulation}
We evaluated the prediction capabilities of the TCN by calculating the mean over 30 different wind gusts included in the training datatset. The rolling mean squared error for the predicted velocity of the blimp using the constant velocity model as well as the TCN-based velocity model are summarized in Table \ref{tab:tcn_cv_comparison}.

\begin{table}[]
\centering
\small
\caption{Rolling MSE of the predictions of the TCN-based velocity model and the constant velocity model, averaged over 30 episodes in the training dataset. \label{tab:tcn_cv_comparison}}
\begin{tabular}{@{} c >{\centering\arraybackslash}p{1cm} >{\centering\arraybackslash}p{2cm}>{\centering\arraybackslash}p{2cm} >{\centering\arraybackslash}p{2cm}@{}}
\toprule
\thead{\small Max. gust\\ \small vel. \\ \small $[\si{m/s}]$} & \thead{\small Dir.\\ \small }   & \thead{\small Avg. RoMSE\\ \small TCN vel. \\ \small $[10^{-4}\si{m^2/s^2}]$} & \thead{\small Avg. RoMSE\\ \small Const. vel. \\ \small $[10^{-4}\si{m^2/s^2}]$}  & \thead{\small Error \\ \small Improv. $[\si{\%}]$}  \\
\midrule
\multirow{3}{*}{1} & x & \textbf{4}  & 7 & 47 \\
                   & y & \textbf{16} & 64 & 75 \\
                   & z & \textbf{12} & 31 & 61 \\
\midrule
\multirow{3}{*}{2} & x & \textbf{16} & 42 & 62 \\
                   & y & \textbf{66} & 268 & 75 \\
                   & z & \textbf{57} & 140 & 59 \\
\midrule
\multirow{3}{*}{3} & x & \textbf{45} & 130 & 65\\
                   & y & \textbf{114} & 420 & 73 \\
                   & z & \textbf{146} & 373 & 61 \\
\midrule
\multirow{3}{*}{4} & x & \textbf{94} & 245 & 62 \\
                   & y & \textbf{233} & 740 & 68 \\
                   & z & \textbf{384} & 798 & 52 \\
\bottomrule
\end{tabular} 
\end{table}

\begin{figure}[htbp]
    \centering
    \includegraphics[width=0.5\textwidth]{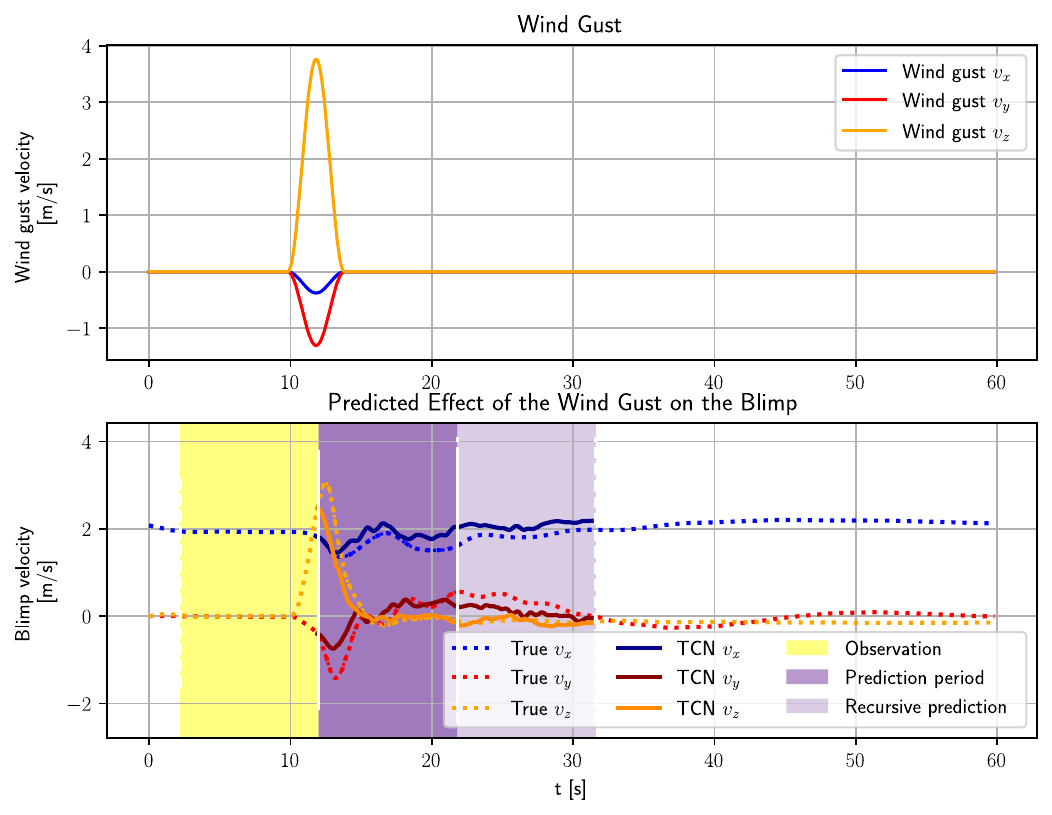}
    \caption{TCN-based prediction of the blimp's velocity when hit by a wind gust with $v_{g,max} = 4\si{m/s}$.}
    \label{fig:wind_gust}
\end{figure}

It can be seen that on average the TCN reduces the velocity prediction error between $47\% $ to $75\%$ over the constant velocity model. This illustrates the \textit{TCN performance}  when predicting the blimp's response to a wind gust. Figure \ref{fig:wind_gust} displays the TCN-based prediction of the blimp's velocity with one additional recursive prediction step. We don't test any higher wind gust velocities than $4m/s$. The reason is that the blimp provided in the airship simulation we use for this work has been shown to fly with a maximum velocity of only $4m/s$ in \cite{price2023}.  Flights under heavier wind conditions are therefore considered too dangerous to attempt docking.

\renewcommand{\arraystretch}{1.6} 

\begin{table*}[h!]
\scriptsize
\setlength{\tabcolsep}{3pt}
\caption{
Docking performance across different scenarios over the 10 episodes in the evaluation dataset. Abbreviation NA: Not available.
All values rounded to nearest integer. 
\label{tab:sim_eval}}
\centering
\begin{tabularx}{\textwidth}{
    c !{\vrule width 0.8pt}
    *{2}{>{\centering\arraybackslash}X}
    !{\vrule width 0.8pt}
    *{3}{>{\centering\arraybackslash}X}
    !{\vrule width 0.8pt}
    *{3}{>{\centering\arraybackslash}X}
}
\Xhline{1pt}
\multirow{2}{*}{\makecell{Eval. \\ Criterion}} & 
\multicolumn{2}{c!{\vrule width 0.8pt}}{CETH Inactive} & 
\multicolumn{3}{c!{\vrule width 0.8pt}}{CETH Active - Const. Vel.} & 
\multicolumn{3}{c}{CETH Active - TCN Vel.} \\
\Xcline{2-9}{0.8pt}
&  No Abort  & Safety Pos. & No Abort & Abort & Safety Pos. & No Abort & Abort & Safety Pos. \\
\Xhline{0.8pt}
Success [\%]  & $0$   & $10$ & $80$ & $100$ & $100$ & $100$ & $100$ & $100$ \\
\Xhline{0.8pt}
\thead{Mean / Std. Dev. \\ Duration [s]}  & NA & $31/0$ & $53/19$ & $60/27$ & $57/19$ & $35/5$ & $46/22$ & $39/8$ \\
\Xhline{0.8pt}
\thead{Min / Max \\ Duration [s]} & NA &  $31/31$ & $32/92$ & $34/134$ & $31/78$ & $27/46$ & $33/109$ & $30/55$ \\
\Xhline{1pt}
\end{tabularx}
\end{table*}

\subsection{Evaluation of Docking Trials in Simulation}\label{sec:evaluation_of_docking_trials}
Table \ref{tab:sim_eval} summarizes the percentage of successful docking trials as well as the mean and standard deviation, minimum and maximum value of the duration required for meeting the docking criteria. In the following, these results are commented in closer detail. 

\subsubsection{General Observations}
Throughout the conduction of experiments it turned out that the complex multi-simulation setup running on two distinct computers had the following issue. 
If the simulation framework ran for a longer time period, the simulated marker-based relative localisation started to fail. This manifested in the situation that the detected position was lagging behind. This led to the situation that the multi-rotor UAV targeted a location slightly behind the docking port. Possible reasons could be the time consuming processing of large 4k images leading to messages being queued inside the ROS framework. For this reason, the simulation framework was restarted frequently, on average every three landing trials to mitigate this issue. 

\subsubsection{CETH Method Inactive} As expected, the scenarios belonging to the category {CA inactive} show the lowest success rate. In fact, only the scenario \textit{Safety position} achieved one single successful docking attempt. This was possible because the UAV traveled in the direction of the safety position first. By chance, this allowed it to return to the docking port on a collision-free trajectory after the wind gust had subsided. 

\subsubsection{CETH Method Active with Constant Velocity Model}
The experiments of the category \textit{CETH Active - Const. Vel.}  allowed success rates of up to $100\%$. 
However, in the scenario \textit{No abort} in the event of a wind gust collisions with the no-fly zone $\mathcal{N}$ occurred. 
This happened during the later stages of the docking approach when the blimp pivoted back to its original straight line trajectory after being deflected by a wind gust. We conclude from this that the constant velocity model was not accurate enough in order to allow precise and collision free maneuvering in the vicinity of the no-fly zone $\mathcal{N}$.
Furthermore, it can be seen that the scenario \textit{ Safety position} requires less time for successful docking than the \textit{Abort}  scenario. This is due to the fact that the safety position leads the multi-rotor UAV away from the tangential hull of the CETH method to a future position of the blimp in the event of a wnd gust. From there approaching the docking port is less time-consuming than from a random position resulting from the UAV being caught in the tangential hull when reacting to pronounced maneuvers of the blimp. Overall, this category of experiments shows the \textit{CETH performance} when the gust detection is active and the docking approach is aborted in the event of a wind gust.

\subsubsection{CETH Method Active with TCN-based Velocity Model}
\begin{figure}[htbp]
    \centering
    \includegraphics[width=0.5\textwidth]{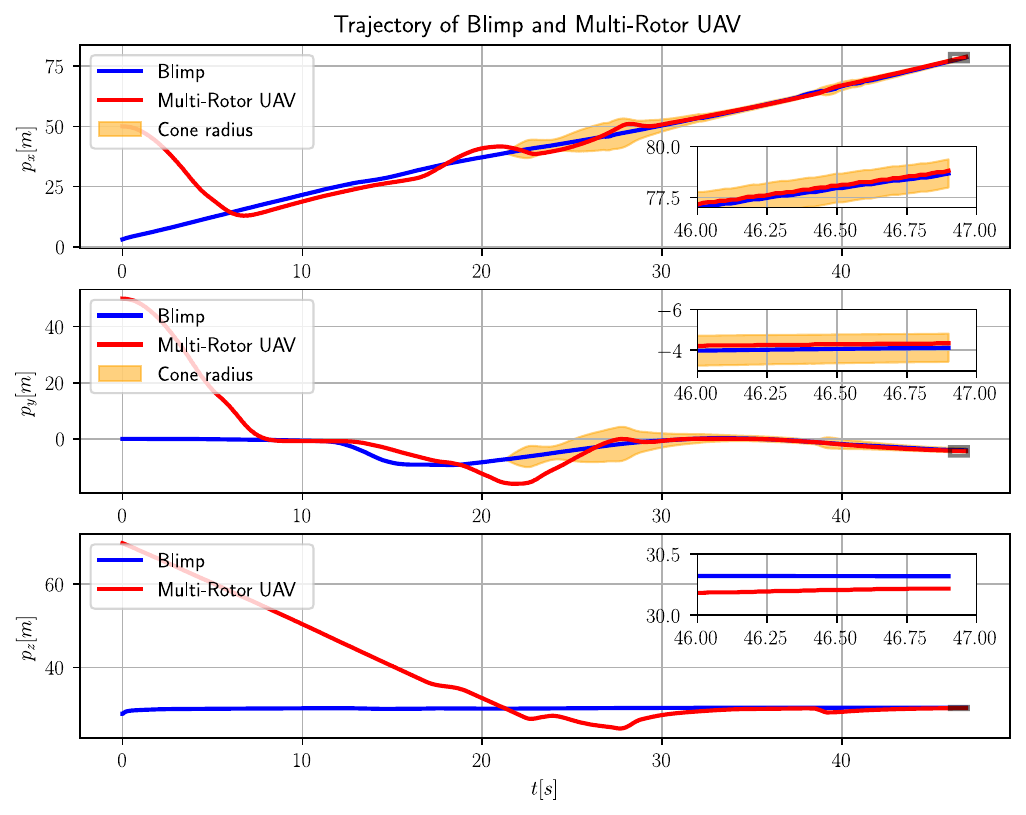}
    \caption{Trajectories of the blimp and multi-rotor UAV throughout an entire docking procedure. Note the varying cone radius of the approach corridor which is displayed depending on the current vertical position of the multi-rotor UAV. }
    \label{fig:trajectories}
\end{figure}

The experiments of the category \textit{CETH Active - TCN Vel.} of the blimp show the best performance. Across all scenarios, no collisions with the no-fly zone $\mathcal{N}$ occurred. 
The benefit of using the TCN-based velocity model over the constant velocity model for the blimp motion is demonstrated in two key results. First, even in the 
\textit{No abort } scenario no collision occurred. This is due to the TCN-based velocity predictions being much more accurate than the constant velocity model (compare Table \ref{tab:tcn_cv_comparison}). This allows to adapt the position of the multi-rotor UAV precisely enough to avoid collision also when flying in close proximity to the no-fly zone $\mathcal{N}$ in the event of a wind gust. Second, the experiments of this category are also the most efficient regarding the duration of the docking procedure. This is another aspect resulting from the accurate maneuvering possibilities facilitating meeting the docking criteria.
Of all experiments of category CETH active - TCN Vel., the scenario  \textit{No abort} has the best performance in terms of time required for docking although it is only slightly better than the scenario \textit{Safety position}.  The reason for the good performance of the latter scenario lies again in the fact that traveling towards the safety position removes the UAV from the tangential hull sending it to a beneficial position from where to begin the docking approach again. However, the advantage of the scenario \textit{Safety position} is that closing the approach corridor in the event of a wind gust adds extra safety to the docking procedure. Figure \ref{fig:trajectories} shows the trajectories of the blimp and the multi-rotor UAV of such a scenario. Note that at the end of the trajectory the multi-rotor UAV  is inside the approach cone and thus in a collision-free location. Overall, this category of experiments illustrates the \textit{TCN benefit} for the CETH method. For this reason, using the CETH method in combination with the safety position is the safest and most efficient docking strategy and therefore the scenario we choose to evaluate in our real world experiments.

\subsection{Evaluation of Docking Trials in the Real World}
\begin{figure}[htbp]
    \centering
    \includegraphics[width=0.5\textwidth]{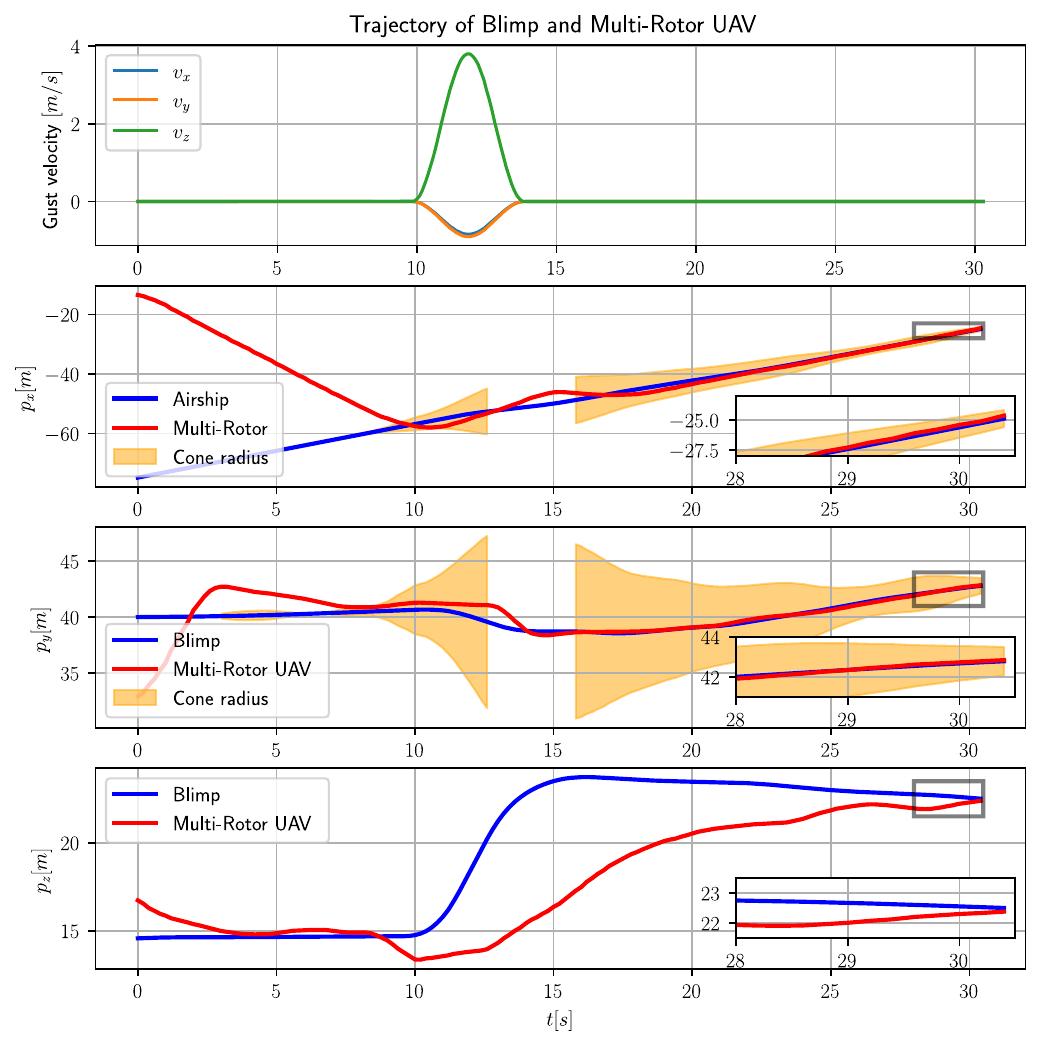}
    \caption{Trajectories of the virtual blimp and the Anafi drone throughout an entire docking procedure recorded in the real world as well as the associated wind gust having an effect on the blimp. Note the gap in the cone radius, here the Anafi drone was not located inside the approach corridor of the CETH method.}
    \label{fig:trajectories_real_world}
\end{figure}

We successfully conducted the docking procedure in three consecutive trials for  a scenario with active CETH method and the safety position. Table~\ref{tab:real_world_exps} shows the starting positions of the blimp and the Anafi drone as well as the time required to meet the docking criteria. The docking procedure was engaged within a time period of $3\si{s}$ after the begin of the blimp trajectory.  
\begin{table}[h!]
\centering
\small
\caption{Parameters of real world experiments. All values rounded to nearest integer}
\label{tab:real_world_exps}
\begin{tabular}{c c c c}
\toprule
\small $~~$Trial$~~$ & $~~$\thead{Initial blimp \\position $[\si{m}]$} $~~$ &$~~$\thead{Initial Anafi \\position $[\si{m}]$} $~~$  & $~~t_{success}~ [\si{s}]~~$ \\
\midrule
1  & $[-74,40,15]^T$ & $[-13,33,17]^T$  & 31 \\
2  & $[-76,40,15]^T$   & $[-15,43,22]^T$ & 42 \\
3  & $[-75,40,15]^T$ &  $[-21,26,10]^T$  & 32 \\

\bottomrule
\end{tabular}
\end{table}
Figure~\ref{fig:trajectories_real_world} illustrates the docking procedure for the second trial. Overall, the real world experiments show a comparable behavior of our docking approach than the simulation results. 
\section{Conclusion} \label{sec:conclusion}
In this work, we proposed a control framework for enabling a multi-rotor UAV to dock on a blimp under the influence of wind gusts. It leverages  a novel TCN-based approach for predicting the response of a blimp to a wind gust. This information is then used as part of an MPC controller planning a collision-free trajectory towards the docking port mounted on the blimp. Collision avoidance is achieved by our novel corridor enhanced tangential hull method allowing maneuvering in close range to the blimp within a designated approach corridor. When a wind gust is detected using our wind gust detection approach, the approach corridor closes and the multi-rotor UAV is steered away. The approach was validated in experiments conducted in simulation and the real world. The results show that the TCN-based prediction of the blimp response significantly outperforms a constant velocity motion model used for the blimp. Experiments in the real world demonstrated that the framework is indeed capable of achieving collision free docking using a virtual blimp trajectory recorded in simulation.
Building on this work, we plan to address the following aspects in future work. First, we intend to extend the dataset with blimp trajectories including non-straight segments such as curves or circles. This would allow us to consider more versatile blimp trajectories during the docking procedure. 
Furthermore, an open challenge remains conducting a docking procedure involving a physical blimp. For this purpose, our current approach is already considering all relevant aspects such as vision-based relative localization.

\printbibliography

@ARTICLE{price2023,
  author={Price, Eric and Black, Michael J. and Ahmad, Aamir},
  journal={IEEE Robotics and Automation Letters}, 
  title={Viewpoint-Driven Formation Control of Airships for Cooperative Target Tracking}, 
  year={2023},
  volume={8},
  number={6},
  pages={3653-3660},
  doi={10.1109/LRA.2023.3264727}}

@ARTICLE{li2003,
  author={Rong Li, X. and Jilkov, V.P.},
  journal={IEEE Transactions on Aerospace and Electronic Systems}, 
  title={Survey of maneuvering target tracking. Part I. Dynamic models}, 
  year={2003},
  volume={39},
  number={4},
  pages={1333-1364},
  doi={10.1109/TAES.2003.1261132}}

@regulation{easa_cs23_amendment_4,
  title={Easy Access Rules for Normal, Utility,
Aerobatic and Commuter Category
Aeroplanes (CS-23) (Amendment 4)},
  organization={EASA},
  number={CS-23},
  version={Amendment 4},
  month={June},
  year={2018},
}

@InProceedings{price2022,
author="Price, Eric
and Liu, Yu Tang
and Black, Michael J.
and Ahmad, Aamir",
editor="Ang Jr, Marcelo H.
and Asama, Hajime
and Lin, Wei
and Foong, Shaohui",
title="Simulation and Control of Deformable Autonomous Airships in Turbulent Wind",
booktitle="Intelligent Autonomous Systems 16",
year="2022",
publisher="Springer International Publishing",
address="Cham",
pages="608--626",
isbn="978-3-030-95892-3"}

@INPROCEEDINGS{tallamraju2018,
  author={Tallamraju, Rahul and Rajappa, Sujit and Black, Michael J. and Karlapalem, Kamalakar and Ahmad, Aamir},
  booktitle={2018 IEEE International Symposium on Safety, Security, and Rescue Robotics (SSRR)}, 
  title={Decentralized MPC based Obstacle Avoidance for Multi-Robot Target Tracking Scenarios}, 
  year={2018},
  volume={},
  number={},
  pages={1-8},
  doi={10.1109/SSRR.2018.8468655}}

@INPROCEEDINGS{nguyen2021,
  author={Nguyen, Huan and Kamel, Mina and Alexis, Kostas and Siegwart, Roland},
  booktitle={2021 European Control Conference (ECC)}, 
  title={Model Predictive Control for Micro Aerial Vehicles: A Survey}, 
  year={2021},
  volume={},
  number={},
  pages={1556-1563},
  doi={10.23919/ECC54610.2021.9654841}}

@ARTICLE{salzmann2023,
  author={Salzmann, Tim and Kaufmann, Elia and Arrizabalaga, Jon and Pavone, Marco and Scaramuzza, Davide and Ryll, Markus},
  journal={IEEE Robotics and Automation Letters}, 
  title={Real-Time Neural MPC: Deep Learning Model Predictive Control for Quadrotors and Agile Robotic Platforms}, 
  year={2023},
  volume={8},
  number={4},
  pages={2397-2404},
  doi={10.1109/LRA.2023.3246839}}

@INPROCEEDINGS{lea2017,
  author={Lea, Colin and Flynn, Michael D. and Vidal, René and Reiter, Austin and Hager, Gregory D.},
  booktitle={2017 IEEE Conference on Computer Vision and Pattern Recognition (CVPR)}, 
  title={Temporal Convolutional Networks for Action Segmentation and Detection}, 
  year={2017},
  volume={},
  number={},
  pages={1003-1012},
  doi={10.1109/CVPR.2017.113}}

@INPROCEEDINGS{liu2022,
  author={Liu, Yu Tang and Price, Eric and Black, Michael J. and Ahmad, Aamir},
  booktitle={2022 IEEE/RSJ International Conference on Intelligent Robots and Systems (IROS)}, 
  title={Deep Residual Reinforcement Learning based Autonomous Blimp Control}, 
  year={2022},
  volume={},
  number={},
  pages={12566-12573},
  doi={10.1109/IROS47612.2022.9981182}}

@article{goldschmid2024,
	author = {Goldschmid, Pascal and Ahmad, Aamir},
	date = {2024/06/06},
	doi = {10.1007/s10514-024-10162-8},
	id = {Goldschmid2024},
	isbn = {1573-7527},
	journal = {Autonomous Robots},
	number = {4},
	pages = {13},
	title = {Reinforcement learning based autonomous multi-rotor landing on moving platforms},
	url = {https://doi.org/10.1007/s10514-024-10162-8},
	volume = {48},
	year = {2024}}

@INPROCEEDINGS{gall2024,
  author={Gall, Christian and Fichter, Walter and Ahmad, Aamir},
  booktitle={2024 IEEE International Conference on Robotics and Automation (ICRA)}, 
  title={End-to-End Thermal Updraft Detection and Estimation for Autonomous Soaring Using Temporal Convolutional Networks}, 
  year={2024},
  volume={},
  number={},
  pages={17875-17881},
  doi={10.1109/ICRA57147.2024.10611479}}

@article{herrera2007,
title = {Recursive prediction for long term time series forecasting using advanced models},
journal = {Neurocomputing},
volume = {70},
number = {16},
pages = {2870-2880},
year = {2007},
note = {Neural Network Applications in Electrical Engineering Selected papers from the 3rd International Work-Conference on Artificial Neural Networks (IWANN 2005)},
issn = {0925-2312},
doi = {https://doi.org/10.1016/j.neucom.2006.04.015},
url = {https://www.sciencedirect.com/science/article/pii/S0925231207001622},
author = {L.J. Herrera and H. Pomares and I. Rojas and A. Guillén and A. Prieto and O. Valenzuela},
}

@INPROCEEDINGS{zhu2016,
  author={Hu Zhu and Suozhong Yuan and Qian Shen},
  booktitle={2016 IEEE Chinese Guidance, Navigation and Control Conference (CGNCC)}, 
  title={Vision/GPS-based docking control for the UAV Autonomous Aerial Refueling}, 
  year={2016},
  volume={},
  number={},
  pages={1211-1215},
  doi={10.1109/CGNCC.2016.7828960}}

@INPROCEEDINGS{pantiskas2020,
  author={Pantiskas, Leonardos and Verstoep, Kees and Bal, Henri},
  booktitle={2020 IEEE Symposium Series on Computational Intelligence (SSCI)}, 
  title={Interpretable Multivariate Time Series Forecasting with Temporal Attention Convolutional Neural Networks}, 
  year={2020},
  volume={},
  number={},
  pages={1687-1694},
  doi={10.1109/SSCI47803.2020.9308570}}

@misc{bai2018,
      title={An Empirical Evaluation of Generic Convolutional and Recurrent Networks for Sequence Modeling}, 
      author={Shaojie Bai and J. Zico Kolter and Vladlen Koltun},
      year={2018},
      eprint={1803.01271},
      archivePrefix={arXiv},
      primaryClass={cs.LG},
      url={https://arxiv.org/abs/1803.01271}, 
}

@article{parsons2019,
author = {Parsons, Christopher and Paulson, Zachary and Nykl, Scott and Dallman, William and Woolley, Brian G. and Pecarina, John},
title = {Analysis of Simulated Imagery for Real-Time Vision-Based Automated Aerial Refueling},
journal = {Journal of Aerospace Information Systems},
volume = {16},
number = {3},
pages = {77-93},
year = {2019},
doi = {10.2514/1.I010658},
URL = {https://doi.org/10.2514/1.I010658},
eprint = {https://doi.org/10.2514/1.I010658},
}

@article{dai2018a,
author = {Dai, Xunhua and Quan, Quan and Ren, Jinrui and Xi, Zhiyu and Cai, Kai-Yuan},
title = {Terminal Iterative Learning Control for Autonomous Aerial Refueling Under Aerodynamic Disturbances},
journal = {Journal of Guidance, Control, and Dynamics},
volume = {41},
number = {7},
pages = {1577-1584},
year = {2018},
doi = {10.2514/1.G003217},
URL = {https://doi.org/10.2514/1.G003217},
eprint = {https://doi.org/10.2514/1.G003217},
}

@article{dai2018b,
title = {Iterative learning control and initial value estimation for probe–drogue autonomous aerial refueling of UAVs},
journal = {Aerospace Science and Technology},
volume = {82-83},
pages = {583-593},
year = {2018},
issn = {1270-9638},
doi = {https://doi.org/10.1016/j.ast.2018.09.034},
url = {https://www.sciencedirect.com/science/article/pii/S1270963818312513},
author = {Xunhua Dai and Quan Quan and Jinrui Ren and Kai-Yuan Cai},
}

@article{liu2019,
title = {Novel docking controller for autonomous aerial refueling with probe direct control and learning-based preview method},
journal = {Aerospace Science and Technology},
volume = {94},
pages = {105403},
year = {2019},
issn = {1270-9638},
doi = {https://doi.org/10.1016/j.ast.2019.105403},
url = {https://www.sciencedirect.com/science/article/pii/S1270963819311496},
author = {Yiheng Liu and Honglun Wang and Jiaxuan Fan},
}

@INPROCEEDINGS{miyazaki2018,
  author={Miyazaki, Ryo and Jiang, Rui and Paul, Hannibal and Ono, Koji and Shimonomura, Kazuhiro},
  booktitle={2018 IEEE/RSJ International Conference on Intelligent Robots and Systems (IROS)}, 
  title={Airborne Docking for Multi-Rotor Aerial Manipulations}, 
  year={2018},
  volume={},
  number={},
  pages={4708-4714},
  doi={10.1109/IROS.2018.8594513}}

@InProceedings{shankar2021,
author="Shankar, Ajay
and Elbaum, Sebastian
and Detweiler, Carrick",
editor="Siciliano, Bruno
and Laschi, Cecilia
and Khatib, Oussama",
title="Multirotor Docking with an Airborne Platform",
booktitle="Experimental Robotics",
year="2021",
publisher="Springer International Publishing",
address="Cham",
pages="47--59",
}

@InProceedings{shankar2024,
author="Shankar, Ajay
and Woo, Heedo
and Prorok, Amanda",
editor="Ang Jr, Marcelo H.
and Khatib, Oussama",
title="Docking Multirotors in Close Proximity Using Learnt Downwash Models",
booktitle="Experimental Robotics",
year="2024",
publisher="Springer Nature Switzerland",
address="Cham",
pages="427--437",
}

@INPROCEEDINGS{caruso2021,
  author={Caruso, Basilio and Fatakdawala, Murtaza and Patil, Akshata and Chen, George and Wilde, Markus},
  booktitle={2021 IEEE Aerospace Conference (50100)}, 
  title={Demonstration of In-Flight Docking Between Quadcopters and Fixed-Wing UAV}, 
  year={2021},
  volume={},
  number={},
  pages={1-9},
  doi={10.1109/AERO50100.2021.9438229}}

@inproceedings{hove2023,
author = {Jonathas Laffita van den Hove and Ewoud J.J. Smeur and Bart D.W. Remes},
editor = {D. Moormann},
title = {Rigid Airborne docking between a fixed-wing UAV and an over-actuated multicopter},
year = {2023},
month = {Sep},
day = {11-15},
booktitle = {14$^{th}$ annual International Micro Air Vehicle Conference and Competition},
address = {Aachen, Germany},
pages = {231--239},
note = {Paper no. IMAV2023-29},
}

@article{zheng2024,
title = {Path planning of stratospheric airship in dynamic wind field based on deep reinforcement learning},
journal = {Aerospace Science and Technology},
volume = {150},
pages = {109173},
year = {2024},
issn = {1270-9638},
doi = {https://doi.org/10.1016/j.ast.2024.109173},
url = {https://www.sciencedirect.com/science/article/pii/S1270963824003067},
author = {Baojin Zheng and Ming Zhu and Xiao Guo and Jiajun Ou and Jiace Yuan},
}

@article{qi2024,
title = {Stratospheric airship trajectory planning in wind field using deep reinforcement learning},
journal = {Advances in Space Research},
year = {2024},
issn = {0273-1177},
doi = {https://doi.org/10.1016/j.asr.2024.08.057},
url = {https://www.sciencedirect.com/science/article/pii/S0273117724008810},
author = {Lele Qi and Xixiang Yang and Fangchao Bai and Xiaolong Deng and Yuelong Pan},
}

@InProceedings{ji2005,
author="Ji, Yongnan
and Hao, Jin
and Reyhani, Nima
and Lendasse, Amaury",
editor="Cabestany, Joan
and Prieto, Alberto
and Sandoval, Francisco",
title="Direct and Recursive Prediction of Time Series Using Mutual Information Selection",
booktitle="Computational Intelligence and Bioinspired Systems",
year="2005",
publisher="Springer Berlin Heidelberg",
address="Berlin, Heidelberg",
pages="1010--1017",
}

@inproceedings{quigley2009,
author="Morgan Quigley and Brian Gerkey and Ken Conley and Josh Faust and
Tully Foote and Jeremy Leibs and Eric Berger and Rob Wheeler and Andrew Ng",
title="ROS: an open-source Robot Operating System",
booktitle="Proc. of the IEEE Intl. Conf. on Robotics and Automation (ICRA)
Workshop on Open Source Robotics",
month = may,
year=2009,
address="Kobe, Japan"
}

@INPROCEEDINGS{goldschmid2025,
  author={Goldschmid, Pascal and Ahmad, Aamir},
  booktitle={2025 European Conference on Mobile Robots (ECMR)}, 
  title={A Multi-Simulation Approach with Model Predictive Control for Anafi Drones}, 
  year={2025},
  volume={},
  number={},
  pages={1-8},
   note = {To appear. Available at \url{https://arxiv.org/abs/2502.10218}}
}
\end{document}